\documentclass[10pt,twocolumn,letterpaper]{article}

\usepackage{cvpr}
\usepackage{times}
\usepackage{helvet}  
\usepackage{courier}  
\usepackage{url}  
\usepackage{epsfig}
\usepackage{graphicx}
\usepackage{amsmath}
\usepackage{amssymb}

\usepackage{color,xcolor}
\usepackage{multirow}
\usepackage{array}
\usepackage{eucal}
\usepackage{cuted}
\usepackage{capt-of}

\definecolor{light-gray}{gray}{0.95}    
\definecolor{orange}{rgb}{1,0.5,0}      
\definecolor{pinkr}{RGB}{255,140,197}
\definecolor{bluer}{RGB}{86,193,255}

\usepackage[colorlinks,linkcolor=red, breaklinks=true,bookmarks=false]{hyperref}
\usepackage[belowskip=-10pt,aboveskip=5pt]{caption}
\cvprfinalcopy 

\def\cvprPaperID{609} 

\begin{document}
	\title{Progressive Pose Attention Transfer for Person Image Generation}
	
{
\author{Zhen~Zhu$^{1}$\thanks{~Equal contribution} , Tengteng~Huang$^{1}$\footnotemark[\value{footnote}] , Baoguang~Shi$^{2}$, Miao~Yu$^1$, Bofei~Wang$^3$, Xiang~Bai$^1$\thanks{~Corresponding author}\\
$^1${\em Huazhong Univ. Sci. and Tech.}, $^2${\em Microsoft, Redmond}, $^3${\em ZTE Corporation}\\ 
{\tt \small \{zzhu, huangtengtng, xbai\}@hust.edu.cn}\\ 
{\tt \small shibaoguang@gmail.com}, {\tt \small zealyu@qq.com}, 
{\tt \small wang.bofei@zte.com.cn}\\
}
}
	
    
	
	\maketitle
	\begin{abstract}
		This paper proposes a new generative adversarial network for pose transfer, i.e., transferring the pose of a given person to a target pose. The generator of the network comprises a sequence of Pose-Attentional Transfer Blocks that each transfers certain regions it attends to, generating the person image progressively. Compared with those in previous works, our generated person images possess better appearance consistency and shape consistency with the input images, thus significantly more realistic-looking.
		The efficacy and efficiency of the proposed network are validated both qualitatively and quantitatively on Market-1501 and DeepFashion.
		Furthermore, the proposed architecture can generate training images for person re-identification, alleviating data insufficiency. Codes and models are available at: \url{https://github.com/tengteng95/Pose-Transfer.git}.
	\end{abstract}

	\section{Introduction}
	\label{introduction}

\begin{figure}
		\centering
		\includegraphics[width=0.48\textwidth]{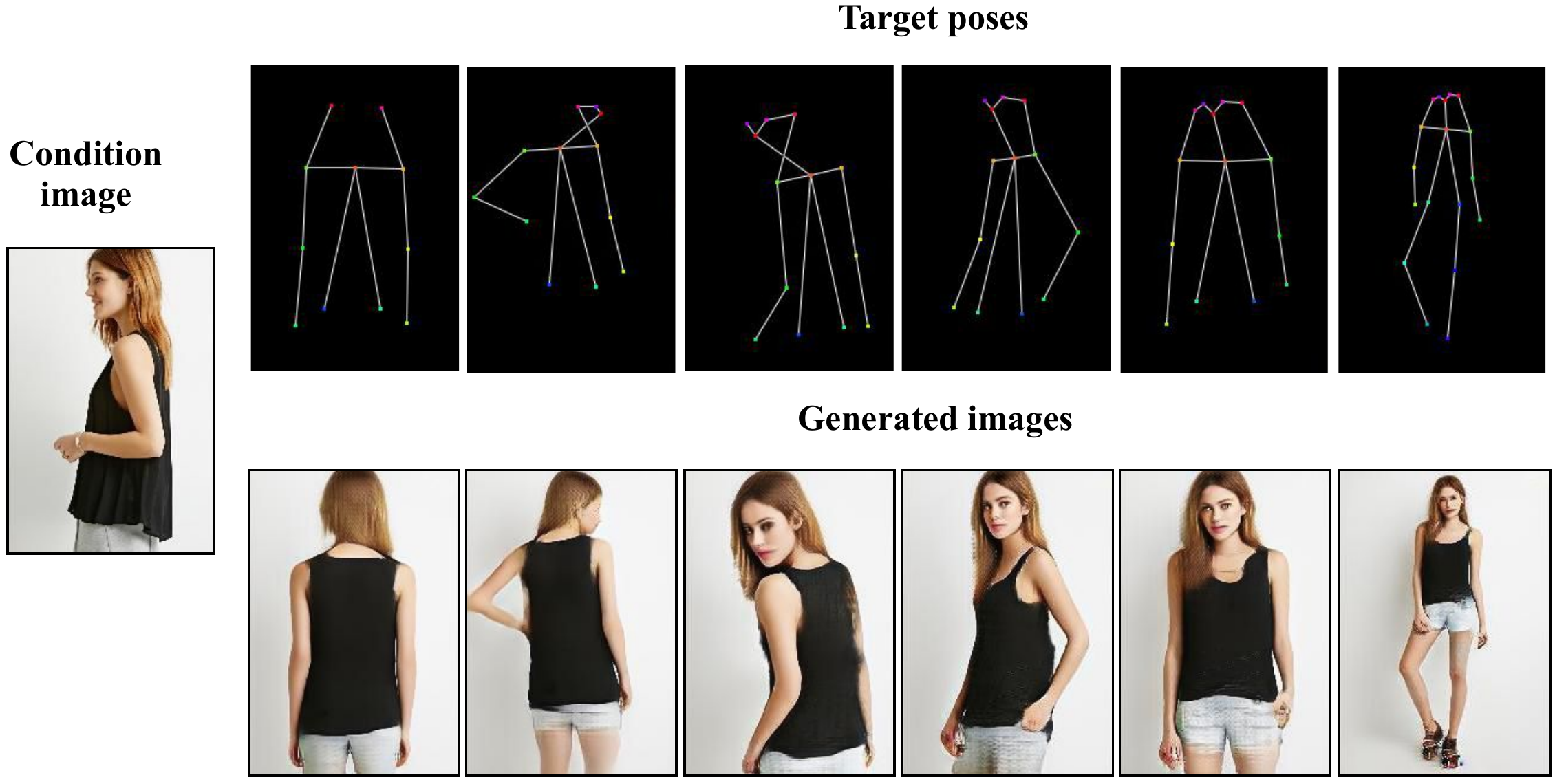}
		\caption{Generated examples by our method based on different target poses. Zoom in for better details. \label{fig:introduction}}
	\end{figure}

	In this paper, we are interested in generating images of non-rigid objects that often possess a large variation of deformation and articulation. 
	Specifically, we focus on transferring a person from one pose to another as depicted in Fig.\ref{fig:introduction}.
	The problem, first introduced by \cite{poseguided} as \emph{pose transfer}, is valuable in many tasks such as video generation with a sequence of poses \cite{videogenerating} and data augmentation for person re-identification \cite{unlabeledReID}. 
	
	Pose transfer can be exceptionally challenging, particularly when given only the partial observation of the person. 
	As exemplified in Fig.~\ref{fig:introduction}, the generator needs to infer the unobserved body parts in order to generate the target poses and views.
	Adding to the challenge, images of different poses under different views can be drastically different in appearance.
	This inevitably demands the generator to capture the large variations possessed by the image distribution.
	Therefore, even with the strong learning capabilities of deep neural networks, recent methods \cite{poseguided,DeformableGAN,Disentangled,vunet} often fail to produce robust results.
	
	We start with the perspective that the images of all the possible poses and views of a certain person constitute a manifold, which is also suggested in previous works \cite{posemanifold2,posemanifold1}. To transfer pose is to go from point $\mathbf{p}_{x}$ on the manifold to another point $\mathbf{p}_{y}$, both indexed by their respective poses.
	In this perspective, the aforementioned challenges can be attributed to the complex structure of the manifold on the global level.
	However, the structure manifold becomes simpler on a local level.
	This transition happens when we restrict the variation of pose to a small range.
	For example, it is hard to transfer pose from sitting to standing, but much simpler to only raise a straight arm to a different angle.
	
	This insight motivates us to take a \emph{progressive} pose transfer scheme.
	In contrast to the one-step transfer scheme adopted in many previous works~\cite{vunet,DeformableGAN}, we propose to transfer a condition pose by transferring through a sequence of intermediate pose representations before reaching the target.
	The transfer is carried out by a sequence of \emph{Pose-Attentional Transfer Blocks} (PATBs), implemented by neural networks.
	This scheme allows each transfer block to perform a local transfer on the manifold, therefore avoiding the challenge of capturing the complex structure of the global manifold.

	\begin{figure*}
		\centering
		\includegraphics[width=0.85\textwidth]{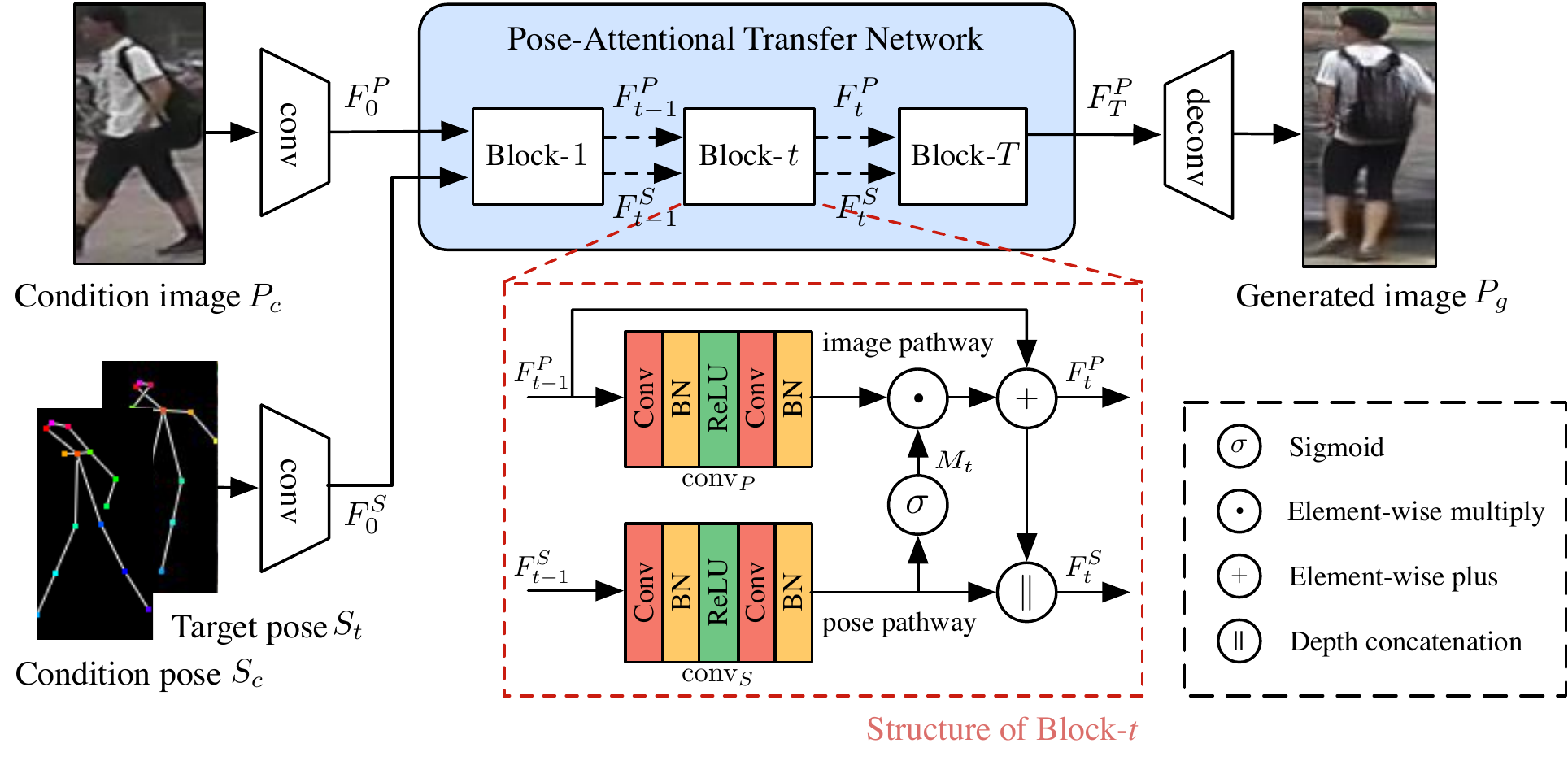}
		\caption{Generator architecture of the proposed method.\label{fig:Generator}}
	\end{figure*}

	Each PATB performs the transfer in a \emph{pose-attentional} manner.
	It takes as input the representation of both the image and the pose.
	Inside the block, we create an attention mechanism that infers the regions of interest based on the human pose.
	As a person's image and pose are registered, this attention mechanism allows better selectivity in choosing the image regions for transferring.
	The block outputs the updated image and pose representations, so that such blocks can be cascaded in sequence to form a PAT network (PATN), as illustrated in Fig.~\ref{fig:Generator}.
	
	The proposed network exhibits superior performance both qualitatively and quantitatively on challenging benchmarks, and substantially augments person dataset for person re-identification application.
	To summarize, the contributions of our paper are two folded:
	
	\begin{enumerate}
		\item We propose a progressive pose attention transfer network to address the challenging task of pose transfer, which is neat in design and efficient in computation.
		\item The proposed network leverages a novel cascaded Pose-Attentional Transfer Blocks (PATBs) that can effectively utilize pose and appearance features to smoothly guide the pose transfer process.
	\end{enumerate}

	\section{Related work}
	Generative Adversarial Networks (GAN) \cite{goodfellow2014generative}, which are basically composed of generator and discriminator that are trained in an adversarial way, can usually generate sharp images~\cite{goodfellow2014generative,cgan,radford2015dcgan,perceptualloss,SRGAN,ACGAN,pix2pix2017-cvpr,CycleGANICCV,ma2018exemplar,yu2018free}. Many real-world applications demand that generated images satisfy some condition constraints, \eg, generating images with specific pose, viewpoint or other attributes. Conditional generative adversarial networks (CGANs) \cite{cgan} are built for this purpose. 
	CGANs have achieved remarkable success in pixel-wise aligned image generation problems. Isola \emph{et al.} \cite{pix2pix2017-cvpr} demonstrated their good applicability for image-to-image translation problems such as day-to-night and sketch-to-image. 
	However, pixel-wise alignment is not well suitable for pose transfer due to the deformation between the condition and target pose. 
	
	For person image generation, Lassner \emph{et al.} \cite{LassnerPG17} presented a model combining VAE \cite{vae} and GAN together to generate images of a person with different clothes, given the 3D model of the person. 
	Zhao \emph{et al.} \cite{multiview} adopted a coarse-to-fine method for generating multi-view cloth images from a single view cloth image. Balakrishnan \emph{et al.} \cite{unseenpose} presented a GAN network that decomposes the person image generation task into foreground and background generation, and then combines to form the final image. 
	Several works \cite{viton,cpviton,humanappearancetransfer,garmenttransfer} were inspired by the virtually try-on applications and made good progress to change the clothes of a given person while maintaining the person pose by warping the clothes to fit for the body topology of the given person.
	
	Specifically for the pose transfer task,
	Ma \emph{et al.} \cite{poseguided} presented a two-stage model to generate person images while their coarse-to-fine strategy requires relatively large computational budget and complicated training procedures. 
	Ma \emph{et al.} \cite{Disentangled} further improved their previous work by disentangling and encoding foreground, background and pose of the input image into embedding features then decodes them back to an image. Though the controllability of the generation process is improved, the quality of their generated images degrade. 
	Likewise, Essner \emph{et al.} \cite{vunet} exploited to combine VAE \cite{vae} and U-Net \cite{pix2pix2017-cvpr} to disentangle appearance and pose. However, appearance features are difficult to be represented by a latent code with fixed length, giving rise to several appearance misalignments.
	Siarohin \emph{et al.} \cite{DeformableGAN} introduced deformable skip connections that require extensive affine transformation computation to deal with pixel-to-pixel misalignment caused by pose differences. And their method is fragile to inaccurate pose disjoints, resulting in poor performance for some rare poses.
	Pumarola \emph{et al.} \cite{unsupervisedposetransfer} adopted a bidirectional strategy to generate person images in a fully unsupervised manner that may induce some geometric errors as pointed out in their paper.
	
	Most of the previous pose transfer approaches adopted keypoint-based pose representation. Besides, Neverova \emph{et al.} \cite{neverova2018dense} adopted \emph{DensePose} \cite{guler2018densepose} as its pose representation, which contains abundant information of depth and body part segmentation, to produce more texture details. The expensive cost of acquiring the DensePose representation for the target pose hinders its applicability, whereas keypoint-based pose representation is much cheaper and more flexible. Hence, we favor to keypoint-based pose representation.

	\section{Model}

\label{sec:network_architecture}
We begin with a few notations. $\left\{ P_{i}^{j}\right\}_{i=1\dots N_{j}}^{j=1\dots M}$
denotes the set of person images in a dataset, where $j$ is person index, $i$ is the image index of person $j$. $M$ is the number of persons, $N_j$ is the number of images of person $j$. 
$S_{i}^{j}$ is the corresponding keypoint-based representation of $P_{i}^{j}$, which consists of a 18-channel heat map that encodes the locations of 18 joints
of a human body. 
We adopt the Human Pose Estimator (HPE) \cite{HPE} used by \cite{poseguided,Disentangled,DeformableGAN} to estimate the 18 joints for fair comparison. During training, the model requires condition and target image $(P_c, P_t)$ and their corresponding condition and target pose heat map $(S_c,S_t)$. The generator outputs a person image, which is challenged by the discriminators for its realness.
The following describes each in detail.


\subsection{Generator}

\subsubsection{Encoders}

Fig.~\ref{fig:Generator} shows the architecture of the generator, whose inputs are the condition image $P_c$, the condition pose $S_c$ and the target pose $S_t$. The generator aims to transfer the pose of the person in the condition image $P_c$ from condition pose $S_c$ to target pose $S_t$, thus generates realistic-looking person image $P_g$.



On the input side, the condition image is encoded by $N$ down-sampling convolutional
layers ($N=2$ in our case); The condition pose $S_c$ and target pose heat maps $S_t$ are stacked along their depth axes before being encoded, also by $N$ down-sampling convolutional
layers. The encoding process mixes the two poses, preserving their
information and capturing their dependencies. Although one can imagine
encoding the two poses separately and concatenate their vectors in
the end, our structure works effectively and requires less computation.

\subsubsection{Pose-Attentional Transfer Network}

At the core of the generator is the \emph{Pose-Attentional Transfer Network} (PATN), consisting of several cascaded Pose-Attentional Transfer Blocks (PATBs). Starting from the initial image code $F_{0}^{P}$ and joint pose code $F_{0}^{S}$, PATN progressively updates these two codes through the sequence of PATBs. At the output, the final image code $F_T^P$ are taken to decode the output image, while the final pose code $F_T^S$ is discarded.

All PATBs have identical structure. A PATB carries out one
step of update. Consider the $t$-th block, whose inputs are $F_{t-1}^{P}$
and $F_{t-1}^{S}$. As depicted in Fig.~\ref{fig:Generator}, the block comprises two pathways, called \emph{image pathway}
and \emph{pose pathway} respectively. With interactions, the two
pathways update $F_{t-1}^{P}$ and $F_{t-1}^{S}$ to $F_{t}^{P}$
and $F_{t}^{S}$, respectively.

In the following, we describe the detailed update process in three
parts and justify their designs.

\vspace{1ex}\noindent\textbf{Pose Attention Masks.}
The pose transfer, at a basic level, is about moving patches from
the locations induced by the condition pose to the locations induced by the target pose. In this sense, the pose guides the
transfer by hinting \emph{where to sample condition patches }and \emph{where to put target patches}. In our network, such hints are realized by the
\emph{attention mask} denoted as $M_t$, which are values between 0 and 1 indicating the importance of every element in the image code.

The attention masks $M_t$ are computed from the pose code $F_{t-1}^{S}$, which incorporates
both the condition and the target pose. The pose code $F_{t-1}^{S}$ firstly goes through
two convolutional layers (with a normalization layer \cite{batchnorm,instancenorm} and ReLU \cite{ReLU} in between), before
being mapped to the range of $(0,1)$ by an element-wise $\mathrm{sigmoid}$
function. Mathematically:
\begin{equation}
\label{w_t}
M_{t}=\sigma\left(\mathrm{conv}_{S}\left(F_{t-1}^{S}\right)\right).
\end{equation} 

\vspace{1ex}\noindent\textbf{Image Code Update.}
Having computed $M_{t}$, the image code $F_{t}^{P}$ is updated by:
\begin{equation}
F_{t}^{P}=M_{t}\odot\text{conv}_{P}\left(F_{t-1}^{P}\right)+F_{t-1}^{P} ,
\end{equation}
where $\odot$ denotes element-wise product. By multiplying the transformed
image codes with the attention masks $M_t$, image code $F_{t}^{P}$ at certain locations
are either preserved or suppressed. The output of the element-wise product is added by $F_{t-1}^{P}$, constituting a \emph{residual connection} \cite{resnet}. 
The residual connection
helps preserve the original image code, which is critical for the
pose transfer. Having residual connection also eases the training as observed in \cite{resnet},
particularly when there are many PATBs.

\vspace{1ex}\noindent\textbf{Pose Code Update.}
As the image code gets updated through the PATBs, the pose code
should also be updated to synchronize the change, \emph{i.e.} update
where to sample and put patches given the new image code. Therefore,
the pose code update should incorporate the new image code.

Specifically, the previous pose code $F_{t-1}^{S}$ firstly goes through two convolutional layers (with a normalization layer and ReLU in between). Note when $t>1$, the first convolutional layer will reduce the feature map depth to half, making the feature map depths of subsequent layers in the pose pathway equal to those in the image pathway.
Then, the transformed code
is mixed with the updated image code by concatenation. Mathematically,
the update is performed by
\begin{equation}
F_{t}^{S}=\mathrm{conv}_{S}\left(F_{t-1}^{S}\right)\Vert F_{t}^{P} ,
\end{equation}
where $\Vert$ denotes the concatenation of two maps along depth
axis.

\subsubsection{Decoder}

The updates in PATN result in the final image code $F_{T}^{P}$ and
the final pose code $F_{T}^{S}$. We take the image code and discard
the final pose code. Following the standard practice, the decoder generates
the output image $P_{g}$ from $F_{T}^{P}$ via $N$ deconvolutional
layers.


\subsection{Discriminators}



We design two discriminators called \emph{appearance discriminator} $D_{A}$ and \emph{shape discriminator} $D_{S}$, to judge how likely $P_{g}$ contains the same person in $P_{c}$ (\emph{appearance consistency}) and how well $P_{g}$ align with the target pose $S_{t}$ (\emph{shape consistency}).
The two discriminators
have similar structures, where $P_{g}$ is concatenated with either
$P_{c}$ or $S_{t}$ along the depth axis, before being fed into a
CNN (convolutional neural network) for judgment. Their outputs are respectively $R^{A}$ and $R^{S}$,
\emph{i.e.} the appearance consistency score and the shape consistency
score. The scores are the probabilities output by the softmax layer
of a CNN. The final score $R$ is the production of the two scores: $R=R^{A}R^{S}$.

As the training proceeds, we observe that a discriminator with low capacity becomes insufficient to differentiate real and fake data. 
Therefore, we build the discriminators by adding three residual blocks after two down-sampling convolutions to enhance their capability. 

\subsection{Training}

	\label{lossfunction}
	The full loss function is denoted as:
	\begin{equation}
	\mathcal{L}_{full}=\arg\underset{G}{\min}\, \underset{D}{\max}\, \alpha \mathcal{L}_{GAN} + \mathcal{L}_{combL1},
	\end{equation}
	where $\mathcal{L}_{GAN}$ denotes the adversarial loss and $\mathcal{L}_{combL1}$ denotes the combined $L_1$ loss. $\alpha$ represents the weight of $\mathcal{L}_{GAN}$ that contributes to $\mathcal{L}_{full}$.
	The total adversarial loss is derived from $D_{A}$ and $D_{S}$:
	
	\begin{small}
		\begin{equation}
		\begin{split}
		\mathcal{L}_{GAN}=&\mathbb{E}_{S_t\in \mathcal{P}_S,(P_c,P_t)\in \mathcal{P}} \left\{\log [D_{A}(P_c,P_t) \cdot D_{S}(S_t,P_t)]\right\} + \\ 
		&\mathbb{E}_{S_t\in \mathcal{P}_S, P_c\in \mathcal{P}, P_g\in \hat{\mathcal{P}}}\left\{\log[(1-D_{A}(P_c,P_g)) \right. \\
		&\left. \cdot (1-D_{S}(S_t,P_g))]\right\}.
		\end{split}
		\end{equation}
	\end{small}
	Note that $\mathcal{P}$, $\hat{\mathcal{P}}$ and $\mathcal{P}_S$ denotes the distribution of real person images, fake person images and person poses, respectively. 
	
	The combined $L_1$ loss can be further written as:
	\begin{equation}
	\label{totalperceptualloss}
	\mathcal{L}_{combL1}=\lambda_1\mathcal{L}_{L1}+\lambda_2\mathcal{L}_{perL1},
	\end{equation}
	where $\mathcal{L}_{L1}$ denotes the pixel-wise $L_1$ loss computed between the generated and target image and $\mathcal{L}_{L1}=\left \| P_{g}-P_t \right \|_1$.

	In order to reduce pose distortions and make the produced images look more natural and smooth, we integrate a \textit{perceptual $L_1$ loss} $\mathcal{L}_{perL1}$, which is effective in super-resolution \cite{SRGAN}, style transfer \cite{perceptualloss} as well as the pose transfer tasks \cite{vunet,DeformableGAN}. 
	In formula, 
	\begin{small}
		\begin{equation}
		\mathcal{L}_{perL1}=\frac{1}{W_{\rho}H_{\rho }C_{\rho}} \sum_{x=1}^{W_{\rho }}\sum_{y=1}^{H_{\rho }}\sum_{z=1}^{C_{\rho }}\left \| \phi _{\rho }(P_{g})_{x,y,z}-\phi _{\rho }(P_t)_{x,y,z} \right \|_1
		\end{equation}
	\end{small} where $\phi _{\rho }$ are the outputs of a layer, indexed by $\rho$, from the VGG-19 model \cite{vgg} pre-trained on ImageNet \cite{ImageNet}, and $W_{\rho}$,$H_{\rho }$,$C_{\rho}$ are the spatial width, height and depth of $\phi _{\rho }$, respectively. We found $\rho = Conv1\_ 2$ leads to the best results in our experiments. 
	
	\vspace{1ex}\noindent\textbf{Training procedures.}
	Following the training procedures of GAN, we alternatively train generator and two discriminators. 
	During training, the generator takes $(P_c, S_c, S_t)$ as input and outputs a transfered person image with target pose $P_{g}$. More specifically, $P_c$ is fed to the image pathway and $(S_c, S_t)$ are fed to the pose pathway. For the adversarial training, $(P_c, P_t)$ and $(P_c, P_{g})$ are fed to the appearance discriminator $D_{A}$ chasing for \textit{appearance consistency}. $(S_t, P_t)$ and $(S_t, P_{g})$ are fed to the shape discriminator $D_{S}$ for \textit{shape consistency}.
	
	\vspace{1ex}\noindent\textbf{Implementation details.}
	The implementation is built upon the popular Pytorch framework.
	Adam optimizer \cite{Kingma2014-Adam} is adopted to train the proposed model for around 90k iterations with $\beta_1=0.5,\beta_2=0.999$. Learning rate is initially set to $2\times 10^{-4}$, and linearly decay to 0 after 60k iterations. We use 9 PATBs in the generator for both datasets. ($\alpha,\lambda_1$, $\lambda_2$) is set to (5, 1, 1) and instance normalization \cite{instancenorm} is applied for DeepFashion and batch normalization \cite{batchnorm} is used  for Market-1501. Dropout \cite{dropout} is only applied in the PATBs, its rate set to 0.5. Leaky ReLU \cite{LeakyReLU} is applied after every convolution or normalization layers in the discriminators, and its negative slope coefficient is set to 0.2.
	
	\section{Experiments}
	In this section, we conduct extensive experiments to verify the efficacy and efficiency of the proposed network. The experiments not only show the superiority of our network but also verify its design rationalities in both objective quantitative scores and subjective visual realness.
	\footnote{Due to page limits, more visual results including a video are put in the supplementary material for further reference.}

	\vspace{1ex}\noindent\textbf{Datasets.} We mainly carry out experiments on the challenging person re-identification dataset Market-1501 \cite{Market1501} and the \emph{In-shop Clothes Retrieval Benchmark} DeepFashion \cite{DeepFashion}. Performing pose transfer on Market-1501 is even challenging since the images are in low-resolution ($128\times64$) and vary enormously in the pose, viewpoint, background and illumination, whereas the images in DeepFashion are in high-resolution ($256\times256$) and of clean backgrounds. We adopt HPE \cite{HPE} as pose joints detector and filter out images where no human body is detected. Consequently, we collect 263,632 training pairs and 12,000 testing pairs for Market-1501. And for DeepFashion, 101,966 pairs are randomly selected for training and 8,570 pairs for testing. 
	It's worth noting that the person identities of the training set do not overlap with those of the testing set for better evaluating the model's generalization ability.
	
	\vspace{1ex}\noindent\textbf{Metrics.}
	It remains an open problem to effectively evaluate the appearance and shape consistency of the generated images. 
	Ma \emph{et al.} \cite{poseguided} used Structure Similarity (SSIM) \cite{SSIM} and Inception score (IS) \cite{improvedtrainingGAN} as their evaluation metrics, then introduced their masked versions -- mask-SSIM and mask-IS, to reduce the background influence by masking it out. Siarohin \emph{et al.} \cite{DeformableGAN} further introduced Detection Score (DS) to measure whether a detector \cite{SSD} can correctly detect the person in the image. We argue that all the metrics mentioned above cannot explicitly quantify the shape consistency. More specifically: SSIM relies on global covariance and means of the images to assess the structure similarity, which is inadequate for measuring shape consistency; IS and DS use image classifier and object detector to assess generated image quality, which are unrelated to shape consistency. 
	
	As all the metrics mentioned above are either insufficient or disable to quantify the shape consistency of the generated images,  we hereby introduce a new metric as a complement to explicitly assess the shape consistency. To be specific, person shape is simply represented by 18 pose joints obtained from the Human Pose Estimator (HPE) \cite{HPE}. Then the shape consistency is approximated by pose joints alignment which is evaluated from PCKh measure\footnote{PCKh is the slightly modified version of Percentage of Correct Keypoints (PCK) \cite{PCK}} \cite{PCKh}. According to the protocol of \cite{PCKh}, PCKh score is the percentage of the keypoints pairs whose offsets are below the half size of the head segment. The head is estimated by the bounding box that tightly covers a set of keypoints related to head.

	\subsection{Comparison with previous work}
	\label{sec:comparison_with_previous_work}
	
	\subsubsection{Quantitative and qualitative comparison}
	\begin{table*}[h]
		\footnotesize
		\centering
		\begin{tabular}{p{2.8cm}<{\centering}|p{0.8cm}<{\centering}p{0.8cm}<{\centering}p{1.5cm}<{\centering}p{1.0cm}<{\centering}p{0.8cm}<{\centering}p{1cm}<{\centering}|p{0.8cm}<{\centering}p{0.7cm}<{\centering}p{0.8cm}<{\centering}c}
			\hline
			\multirow{2}{*}{Model} & \multicolumn{6}{c|}{Market-1501}           & \multicolumn{4}{c}{DeepFashion} \\ \cline{2-11} 
			& SSIM  & IS    & mask-SSIM & mask-IS & DS  & PCKh & SSIM      & IS       & DS    & PCKh   \\ \hline
			Ma \emph{et al.} \cite{poseguided}  & 0.253 & 3.460          & 0.792     & 3.435   & -     & -     & 0.762    & 3.090   & -   & -     \\
			Ma \emph{et al.} \cite{Disentangled}    & 0.099 & 3.483 & 0.614     & 3.491   & -     & -     & 0.614    & 3.228    & -       & -     \\
			Siarohin \emph{et al.} \cite{DeformableGAN}           & 0.290 & 3.185          & 0.805     & 3.502   & 0.72  & -     & 0.756  & 3.439    & 0.96   & -     \\  \hline
			Ma \emph{et al.} * \cite{poseguided} & 0.261 & \textbf{3.495} & 0.782     & 3.367   & 0.39  & 0.73  & 0.773    & 3.163    & 0.951   & 0.89  \\
			Siarohin \emph{et al.} * \cite{DeformableGAN}          & 0.291 & 3.230          & 0.807     & 3.502   & 0.72  & 0.94  & 0.760    & 3.362    & 0.967   & 0.94   \\ 
			Esser \emph{et al.} * \cite{vunet}          & 0.266 & 2.965 & 0.793 & 3.549 & 0.72 & 0.92 & 0.763 & \textbf{3.440} & 0.972 & 0.93 \\
			Ours             & \textbf{0.311}  & 3.323  & \textbf{0.811}  & \textbf{3.773}  & \textbf{0.74}   & \textbf{0.94}  & \textbf{0.773}  & 3.209 & \textbf{0.976}  & \textbf{0.96} \\  \hline
			Real Data  & 1.000 & 3.890  & 1.000  & 3.706  & 0.74  & 1.00   & 1.000  & 4.053    & 0.968 & 1.00        \\  \hline
		\end{tabular}
		\caption{Comparison with state-of-the-art on Market-1501 and DeepFashion. * denotes the results tested on our test set.}
		\label{quantitative-comparison}
	\end{table*}

	Quantitative comparisons with previous works can be found in Tab.\ref{quantitative-comparison}. 
	Since the data split of previous works \cite{poseguided,DeformableGAN} are not given, we download their well-trained models and re-evaluate their performance on our testing set.
	Although the testing set inevitably includes some of their training images and thus giving an edge to their methods, our method still outperforms them on most metrics, and the numeric improvements are steady for both datasets. Notably, our method ranks the highest on the PCKh metric which captures shape consistency. Although we achieve the same highest PCKh score as \cite{DeformableGAN} on Market-1501, we promote the best PCKh performance of previous works by 2\% on DeepFashion, attaining to 0.96, which nearly hits the performance ceiling. 
	
	\begin{figure}[h]
		\centering
		\includegraphics[width=0.48\textwidth]{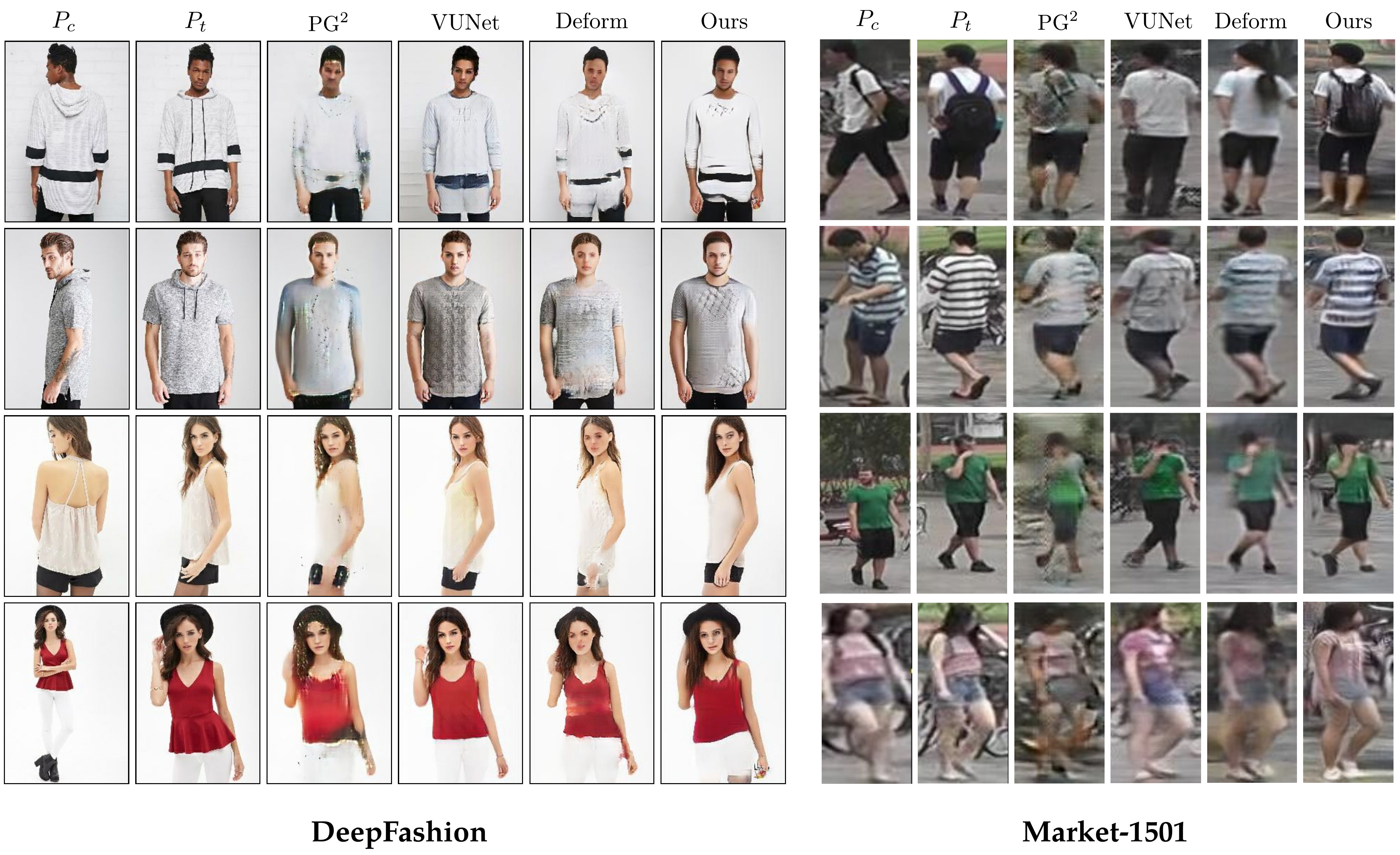}
		\caption{Qualitative comparisons on DeepFashion (Left) and Market-1501 (Right) dataset. $\mathrm{PG^2}$, $\mathrm{VUnet}$ and $\mathrm{Deform}$ represent the results of \cite{poseguided}, \cite{vunet} and \cite{DeformableGAN}, respectively.}
		\label{fig:qualitative_comparison}
	\end{figure}
	

	Fig.\ref{fig:qualitative_comparison} (Left) gives some typical qualitative examples, all of which are with large pose changes and/or scale variance, on the high-resolution DeepFashion \cite{DeepFashion}. Overall, our proposed method maintains the integrity of the person, especially on the wrist, and exhibits the most natural posture. Meanwhile, our method captures the most detailed appearance, especially the skin color in the first row, whiskers in the second row, hair in the third row and hat in the last row. Besides, our method generates much more beautiful facial details than other methods.

	
    We also evaluate the performance of our method on Market-1501, a dataset of poor image quality. Some examples are shown in Fig.\ref{fig:qualitative_comparison} (Right). Our method yields the sharpest person images while the generated images of other methods are blurred to some extent. Especially, our method gives the correct leg layouts that correspond to the target poses, even when the legs are crossed in the target pose (in the second and third row), or the condition image is blurred (in the last row). Moreover, we achieved the best appearance consistency, \eg, the bag is presented in our result while lost by other methods (in the first row).
	
	\subsubsection{Model and computation complexity comparison}
    Tab.\ref{tab:efficiency} gives the model and computation complexity of all the three methods.
	These methods are tested under one NVIDIA Titan Xp graphics card in the same workstation. Only GPU time is taken into account when generating all the testing pairs of DeepFashion to compute the speed.
	Notably, our method significantly outperforms other methods in both the number of parameters and the computation complexity, owing to the simple and neat structure of the building blocks (PATBs) of our network. In contrast, the two-stage strategy of Ma \emph{et al.} \cite{poseguided} brings a huge increase in parameters and computational burden. Although Siarohin \emph{et al.} \cite{DeformableGAN} switched to one stage model by introducing the deformable skip connections (DSCs) leading to a decreased model size, the computation-intensive body part affine transformations required by DSCs make the decrease of computation complexity only marginal. 
	
	\begin{table}[h]
		\centering
		\footnotesize
		\begin{tabular}{c|c|c}
			\hline
			Method          & Params  & Speed  \\ \hline
			Ma \emph{et al.} \cite{poseguided}       & 437.09 M       &  10.36 fps   \\
			Siarohin \emph{et al.} \cite{DeformableGAN} & 82.08 M        &  17.74 fps     \\
			Esser \emph{et al.} \cite{vunet} & 139.36 M & 29.37 fps\\
			Ours (9 PATBs)            & 41.36 M        &  60.61 fps     \\ \hline
		\end{tabular}
		\caption{Comparison of model size and testing speed on DeepFashion dataset. ``M'' denotes millions and ``fps'' denotes Frames Per Second. \label{tab:efficiency} }
	\end{table}

	\subsubsection{User study}
	Generally, it is more appropriate to judge the realness of the generated images by human. We recruit 30 volunteers to give an instant judgment (real/fake) about each image within a second. Following the protocol of \cite{poseguided,DeformableGAN}, 55 real and 55 generated images are randomly selected and shuffled, the first 10 of them are used for practice then the remaining 100 images are used as the evaluation set. Our method makes considerable improvements over \cite{poseguided,DeformableGAN} on all measurements, as shown in Tab.\ref{tab:user_study}, further validating that our generated images are more realistic, natural and sharp. It's worth noticing that our method quite excels at handling condition images of poor quality, where 63.47\% of our generated images are regarded as real by volunteers, reflected as the ``G2R'' measure in Tab.\ref{tab:user_study}.

	\begin{table}[h]
		\centering
		\footnotesize
		\begin{tabular}{l|p{1cm}<{\centering}p{0.8cm}<{\centering}|p{1cm}<{\centering}p{0.8cm}<{\centering}}
			\hline
			\multirow{2}{*}{Model} & \multicolumn{2}{c|}{Market-1501}           & \multicolumn{2}{c}{DeepFashion} \\ \cline{2-5}  & R2G & G2R & R2G & G2R\\ \hline
			Ma \emph{et al.} \cite{poseguided} & 11.2 & 5.5 & 9.2 & 14.9 \\
			Siarohin \emph{et al.} \cite{DeformableGAN} & 22.67 & 50.24 & 12.42 & 24.61 \\
			Ours & 32.23 & 63.47 & 19.14 & 31.78 \\ \hline
		\end{tabular}
		\caption{User study ($\%$). \textbf{\emph{R2G}} means the percentage of real images rated as generated w.r.t. all real images. \textbf{\emph{G2R}} means the percentage of generated images rated as real w.r.t. all generated images. The results of other methods are drawn from their papers.}
		\label{tab:user_study}
	\end{table}

	\subsection{Ablation study and result analysis}
	\label{sec:ablation_study}
	\begin{figure}[h]
		\centering
		\includegraphics[width=0.48\textwidth]{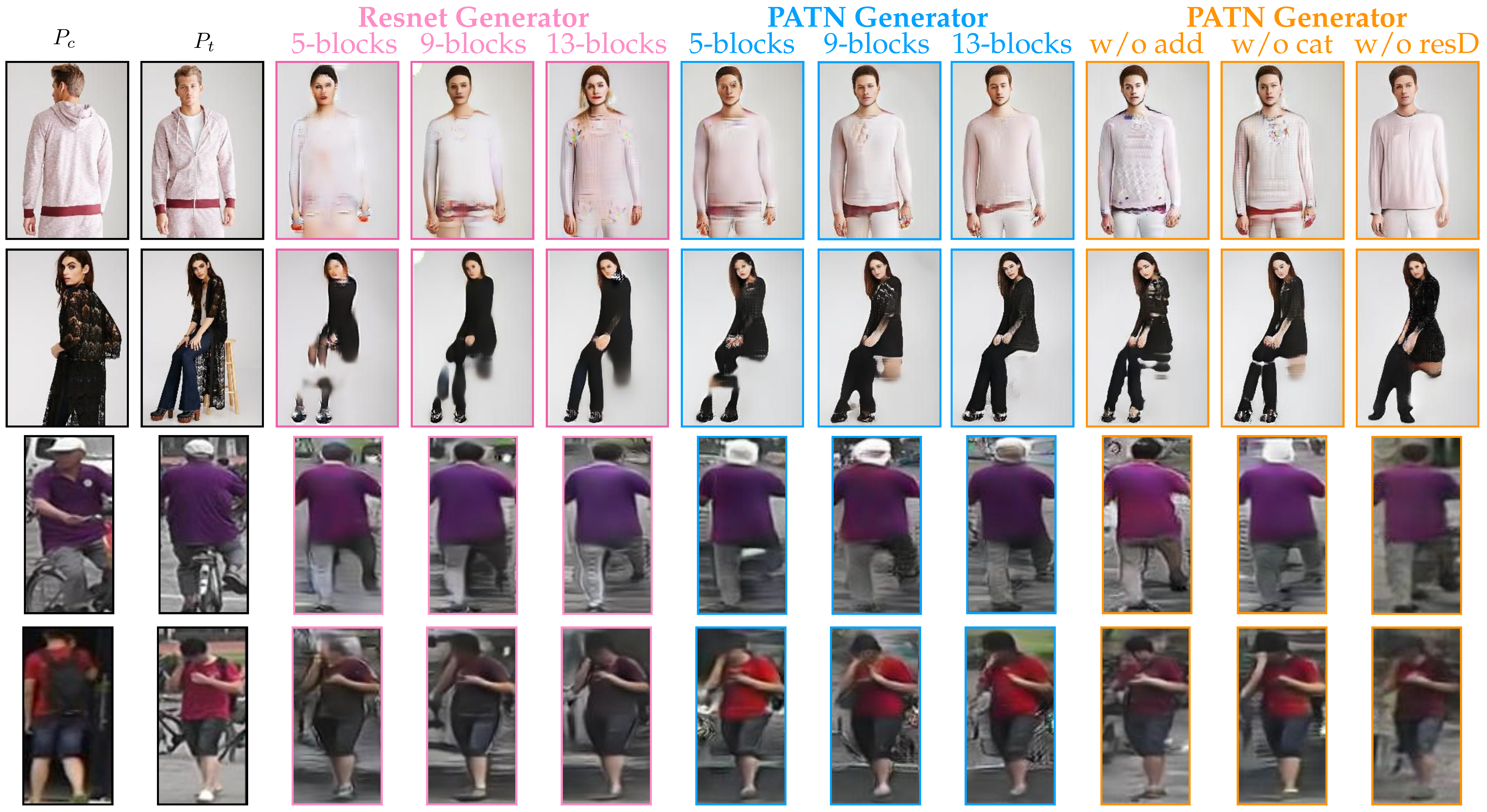}
		\caption{Qualitative results of the ablation study on Market-1501 and DeepFashion dataset. The images with pink border are generated by Resnet generator \cite{perceptualloss,pix2pix2017-cvpr} containing 5, 9, 13 residual blocks \cite{resnet} from left to right respectively. And the images with blue borders are generated by our \emph{PATN generator} containing 5, 9, 13 PATBs. The images with orange borders are generated by discarding a certain part of the whole model.}
		\label{fig:ablation_study}
	\end{figure}

	\vspace{1ex}\noindent\textbf{Efficacy of the main contributions.}~The generator of our network -- PATN has two important design characteristics: one is the carefully designed building block -- PATB which aims to optimize appearance and pose simultaneously using the attention mechanism, and the other is the cascade of building blocks which aims to guide the deformable transfer process progressively. Therefore, we carry out two comparison experiments, one is to explore the advantage of PATB by replacing it with vanilla residual block \cite{perceptualloss} that results in a generator named \emph{Resnet generator}, and the other is to exhibit the advantage of the progressive manner by varying the number of PATBs.
	
	Qualitative comparison results are shown in Fig.\ref{fig:ablation_study}. Comparing the images yield by Resnet generator and those yield by our PATN generator with the same number of building blocks, it is evident that PATN generator can always generate images demonstrating much more consistent shape and appearance with its target image. 
	
	Moreover, the Resnet generator is prone to ignore some indistinguishable but representative appearance information which usually occupies a small portion of the image, and fails to generate correct foreground shape when the target pose is relatively rare. For instance, Resnet generator is likely to ignore the bottom red ring of the sweater in the first row, the white cap in the third row and mislead the T-shirt color as black in the fourth row since a black backpack occludes a large portion of the T-shirt. Besides, the shapes of the sitting girls produced by Resnet generator in the second row are somewhat incomplete as this pose is quite rare in DeepFashion \cite{DeepFashion}. 
	
	As a contrast, our PATN generator with only 5 PATBs could do better than the Resnet generator with 13 residual blocks. We assume this should be attributed to the pose attention mechanism which enhances model's abilities in capturing useful features and leveraging them. And our PATN can produce much finer and more pleasant person images with 9 PATBs. Further increasing the number of PATBs to 13 can marginally boost the performance. This phenomenon reflects that increasing the number of PATBs will ease the transfer process and give results with less artifacts. For a tradeoff, we chose 9 PATBs as default for better efficiency. More qualitative results are given in the supplementary materials for further reference.
	
	Under almost all the quantitative measures shown in Tab.\ref{ablationstudy_v2}, our PATN generator with only 5 PATBs outperforms Resnet generator with all its number of residual blocks configurations. These results clearly demonstrate the advantages of our PATN generator.
	
	\vspace{1ex}\noindent\textbf{Dissection of the PATB operations and the discriminator modifications.}~Besides, to investigate the effect of each part of PATB, we conduct experiments by removing the addition (\emph{w/o add}) and concatenation (\emph{w/o cat}) operation in every PATB inside the PATN generator (9 blocks). To discuss how many improvements obtained from the modified discriminators, we further added an experiment by discarding the residual blocks in discriminators (\emph{w/o resD}). The qualitative and quantitative results are also given in Fig.\ref{fig:ablation_study} and Tab.\ref{ablationstudy_v2}. It's shown that by removing any parts of PATB would lead to a performance drop, visually exhibiting a certain degree of color distortion and implausible details. Discarding residual blocks in the discriminators also reduces local details and influences the integrity of the person body.
	
	\begin{table*}[h]
		\centering
		\footnotesize
		\begin{tabular}{p{3.5cm}<{\centering}|p{1cm}<{\centering}p{0.8cm}<{\centering}p{1.4cm}<{\centering}p{1.3cm}<{\centering}p{0.6cm}<{\centering}p{0.8cm}<{\centering}|p{1cm}<{\centering}p{0.8cm}<{\centering}p{0.6cm}<{\centering}p{1cm}<{\centering}}
			\hline
			\multirow{2}{*}{Model} & \multicolumn{6}{c|}{Market-1501}           & \multicolumn{4}{c}{DeepFashion} \\ \cline{2-11}  & SSIM & IS & mask-SSIM & mask-IS & DS &PCKh &SSIM &IS &DS &PCKh \\ \hline
			Resnet generator (5 blocks) & 0.297 & 3.236 & 0.802 & 3.807 & 0.67 & 0.84 & 0.764 & 2.952 & 0.900 &0.89 \\ 
			Resnet generator (9 blocks)     & 0.301 & 3.077 & 0.802 & 3.862 & 0.69 & 0.87 & 0.767 &3.157 & 0.941 &0.90 \\ 
			Resnet generator  (13 blocks) & 0.300 & 3.134 &0.797 & 3.731 & 0.67 &0.88 &0.766 &3.107 & 0.943 &0.89  \\ \hline \hline
			PATN generator (5 blocks)     & 0.309 & 3.273 & 0.809 &3.870 & 0.69 & 0.91 &0.771 &3.108 & 0.958 &0.93 \\
			PATN generator (9 blocks)  &0.311 & 3.323 & 0.811 & 3.773 & 0.74 & 0.94 &0.773  & 3.209 & 0.976  & 0.96    \\
			PATN generator (13 blocks) & 0.314 & 3.274 &0.808 & 3.797 & 0.75 &0.93 & 0.776 &3.440 & 0.970 & 0.96  \\ \hline \hline
		    PATN generator w/o add  & 0.305 & 3.650 & 0.805 & 3.899 & 0.73 & 0.92 & 0.769 & 3.265 & 0.970 & 0.93  \\
		    PATN generator w/o cat  & 0.306 & 3.373 & 0.807 & 3.915 & 0.71 & 0.92 & 0.767 & 3.200 & 0.968 & 0.95  \\
		    PATN generator w/o resD  & 0.308 & 3.536 & 0.809 & 3.606 & 0.73 &0.93 & 0.773 & 3.442 & 0.978 & 0.94  \\ \hline
		\end{tabular}
		\caption{Quantitative comparison of the ablation study.}
		\label{ablationstudy_v2}
	\end{table*}

\begin{figure}[h]
		\centering
		\small
		\includegraphics[width=0.48\textwidth]{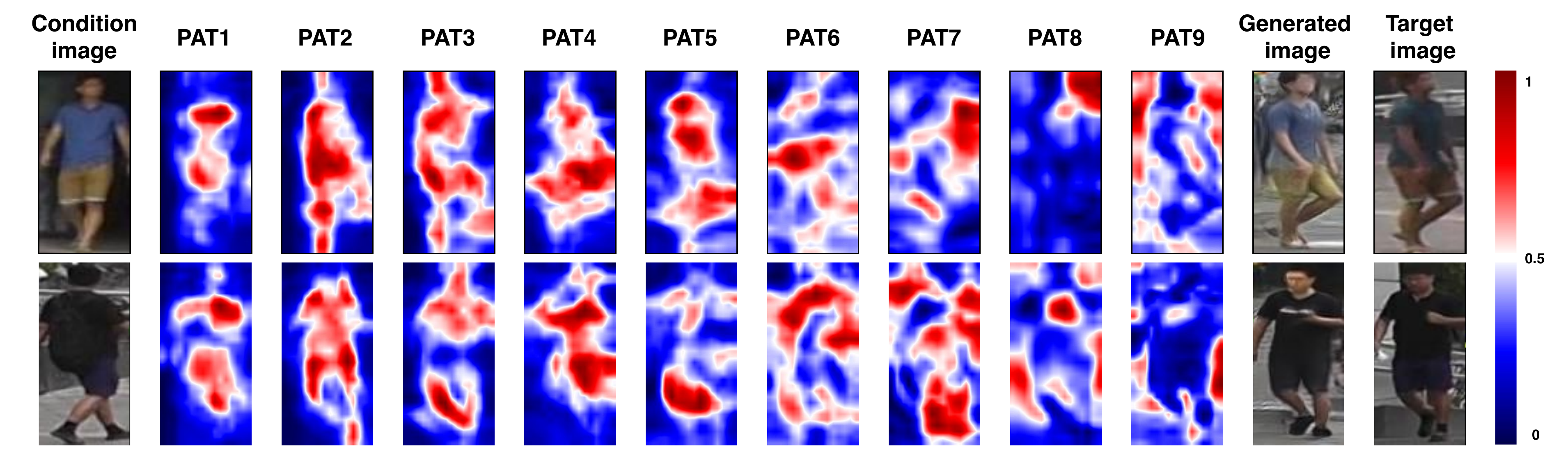}
		\caption{Visualization of the attention masks in PATBs. ``PAT1'' denotes the attention mask of the first PATB and ``PAT2$\sim$9'' likewise.} 
		\label{fig:attblock_vis}
	\end{figure}

	\vspace{1ex}\noindent\textbf{Visualization of the attention masks in PATBs.}~To get an intuitive understanding on how the PATBs work during the pose transfer process, we visualize the attention masks of all the nine cascaded PATBs in Fig.\ref{fig:attblock_vis}. The pose attention transfer process is clear and interpretable, where the attention masks always attend to the regions needed to be adjusted in a progressive manner. The initial several masks (first three columns of masks) are some sort of blending of condition pose and target pose. As the condition pose is transferred towards the target pose, the regions needed to be adjusted are shrunk and scattered, reflected in the middle four mask columns. Eventually, the last two mask columns show that the attention is turned from foreground to background for refinement. 
	
	\section{Application to person re-identification}
	
	A good person pose transfer method can generate realistic-looking person images to augment the datasets of person-related vision tasks, which might result in better performance especially in the situation of insufficient training data. As person re-identification (re-ID) \cite{personreid} is increasingly promising in real-world applications and has great research significance, we choose to evaluate the performance of our method in augmenting the person re-ID dataset. Specifically, we use the mainstream person re-ID dataset Market-1501 \cite{Market1501} and two advanced backbone network, ResNet-50 \cite{resnet} and Inception-v2 \cite{inception}, as our test bed.
	

    We first randomly select a portion $p$ of the real Market-1501 dataset as the reduced training set, denoted as $\mathcal{M}^{R}_{p}$, where at least one image per identity is preserved in $\mathcal{M}^R_{p}$ for better person identity variety. Each unselected training image $\widetilde{I}_i$ is replaced by a generated image that transfers from a randomly chosen image with the same identity in $\mathcal{M}^R_{p}$ to the pose of $\widetilde{I}_i$.
	Consequently, an augmented training set $\mathcal{M}^A_{p}$ is formed from all these generated images and $\mathcal{M}_{p}^R$. 
	The selected portion $p$ varies from 10\% to 90\% at intervals of 10\%. We follow the training and testing protocols of the Open-Reid Framework\footnote{\url{https://github.com/Cysu/open-reid}} to train the ResNet-50 and Inception-v2 on each of the nine reduced training set $\mathcal{M}^R_{p}$  and the nine augmented training set $\mathcal{M}^A_{p}$ respectively. 
	
	
	\begin{table}[h]
	\centering
	\tiny
	\begin{tabular}{p{0.7cm}<{\centering}|p{0.2cm}<{\centering}p{0.2cm}<{\centering}p{0.2cm}<{\centering}p{0.2cm}<{\centering}p{0.2cm}<{\centering}p{0.2cm}<{\centering}p{0.2cm}<{\centering}p{0.2cm}<{\centering}p{0.2cm}<{\centering}p{0.2cm}<{\centering}|p{0.25cm}<{\centering}p{0.25cm}<{\centering}}
		\hline
		\multirow{2}{*}{Aug. Model} & \multicolumn{10}{c|}{Portion $p$ of the real images} & \multicolumn{2}{c}{Aug. ratio}\\ 
		\cline{2-13}  &0.1 &0.2 &0.3 &0.4 &0.5 &0.6 &0.7 &0.8 &0.9 &1.0 &2 &3 \\\hline
        None~(\textbf{Inc}) &5.7 &16.6 &26.0 &33.5 &39.2 &42.2 &46.5 &49.0 &49.5 &52.7 &- &- \\
		Ours &42.3 &43.6 &45.7 &46.7 &48.0 &48.5 &49.1 &50.7 &51.6 &52.7 &56.6 &57.1 \\\hline
        None~(\textbf{Res})     &9.2 &27.6 &41.5 &50.3 &56.2 &58.8 &61.2 &62.7 &63.8 &65.3 &- &- \\ 
		VUNet\cite{vunet}  &49.8 &51.7 &53.7 &54.5 &56.6 &58.4 &59.4 &61.3 &62.9 &65.3 &63.9 &64.1\\ 
		Deform\cite{DeformableGAN} &51.9 &53.9 &55.4 &56.1 &57.6 &59.4 &60.5 &62.2 &63.5 &65.3 &64.2 &64.6 \\
		Ours     &52.6 &54.5 &\textbf{56.5} &56.6 &\textbf{60.3} &\textbf{60.9} &\textbf{62.1} &\textbf{63.3} &\textbf{64.8} &65.3 &\textbf{65.3} &\textbf{65.7} \\
		Ours*     &\textbf{53.3} &\textbf{55.9} &56.0 &\textbf{57.3} &58.8 &60.4 &60.7 &63.1 &64.5 &65.3 &65.1 &65.4 \\ \hline
		
	\end{tabular}
	\caption{The ReID results on Inception-v2~(denoted by \textbf{Inc}) and ResNet-50~(denoted by \textbf{Res}) using images generated by different methods. \textit{None} means no generative model is employed. * denotes the results when we randomly select target poses from $M^R_p$ for data augmentation.}
	
	\label{tab:reid_aug}
    \end{table}
	
	We argue that our experiment scheme is well-suited for assessing the usefulness of our image generation method in boosting the performance of re-ID via data augmentation due to the following two reasons: 1) the original training set reduction leads to a degree of data insufficiency, offering an opportunity to boost the performance via data augmentation and giving the performance \emph{lower bound}; 2) the reference results utilize all the real images including all the ground truth of the generated images, thus could give the theoretical performance \emph{upper bound}. Besides, the gaps between the upper and lower bounds could measure the maximum potential of data augmentation in boosting re-ID performance. 
	
	Tab.\ref{tab:reid_aug} shows that whenever there's a performance gap, there is undoubtedly a performance boost owing to our data augmentation. And the performance gain is much more significant if the performance gap is relatively large. A natural question is, what the performance would be when further augmenting the whole real training set. To investigate, we augment the training dataset of Market-1501 by generating one/two samples per image, whose target poses are randomly selected from the whole dataset, thus doubles/triples the size of the original training set. The results on these two augmented datasets are added to Tab.\ref{tab:reid_aug} (the right part). In a nutshell, the trends of performance gain by adding more generated images are roughly in accordance with those by adding more real images. It can be seen that the real data augmentation for Inception-v2 model could get nearly linear improvements even in cases of near sizes of the whole real data set. Therefore, doubling or tripling the training set continuously improves the performance considerably. On the other hand, real data augmentation for ResNet-50 model tends to saturate in cases of near sizes of the whole real data set, hence doubling or tripling the size fails to improve the performance further.
	
	We further compare our method to several existing person image generators~\cite{vunet, DeformableGAN} under the same setting for re-ID data augmentation. As shown in Tab.\ref{tab:reid_aug}, our method achieve consistent improvements over previous works for different portion $p$ of the real images, suggesting the proposed method can generate more realistic human images and be more effective for the re-ID task. We also present the re-ID performance by randomly selecting the target poses from the whole dataset, which can better suit the practical applications. As shown in Tab.\ref{tab:reid_aug}, there is not an obvious performance difference between the two settings, further demonstrating that the proposed framework is robust to various poses.

	\section{Conclusion}
	In this paper, we propose a progressive pose attention transfer network to deal with the challenging pose transfer. The network cascades several Pose Attentional Transfer Blocks (PATBs), each is capable of optimizing appearance and pose simultaneously using the attention mechanism, thus guides the deformable transfer process progressively. Compared to previous works, our network exhibits superior performance in both subjective visual realness and objective quantitative scores while at the same time improves the computational efficiency and reduces the model complexity significantly. 
	Moreover, the proposed network can be used to alleviate the insufficient training data problem for person re-identification substantially. 
	Additionally, our progressive pose-attentional transfer process can be easily visualized by its attention masks, making our network more interpretable. Moreover, the design of our network has been experimentally verified through the ablation studies.

	Our progressive pose-attentional transfer network is not only specific to generating person images but also can be potentially adapted to generate other non-rigid objects. 
	Furthermore, we assume the idea behind our progressive attention transfer approach may be beneficial to other GAN-based image generation approaches as well. 
	
	\section*{Acknowledgement}
	This work was supported by NSFC 61573160, to Dr. Xiang Bai by the National Program for Support of Top-notch Young Professionals and the Program for HUST Academic Frontier Youth Team. We thank Patrick Esser for his kind instructions on testing VUnet. We would like to express our sincere gratitudes to Jiajia Chu for her valuable help to this paper.
	
	{\small
		\bibliographystyle{ieee}
		\bibliography{egbib}

\begin{thebibliography}{10}\itemsep=-1pt

\bibitem{PCKh}
Mykhaylo Andriluka, Leonid Pishchulin, Peter Gehler, and Bernt Schiele.
\newblock 2d human pose estimation: New benchmark and state of the art
  analysis.
\newblock In {\em Proc. CVPR}, 2014.

\bibitem{unseenpose}
Guha Balakrishnan, Amy Zhao, Adrian~V. Dalca, Fr{\'{e}}do Durand, and John~V.
  Guttag.
\newblock Synthesizing images of humans in unseen poses.
\newblock {\em CoRR}, abs/1804.07739, 2018.

\bibitem{HPE}
Zhe Cao, Tomas Simon, Shih{-}En Wei, and Yaser Sheikh.
\newblock Realtime multi-person 2d pose estimation using part affinity fields.
\newblock In {\em Proc. CVPR}, pages 1302--1310, 2017.

\bibitem{posemanifold2}
Ahmed~M. Elgammal and Chan{-}Su Lee.
\newblock Inferring 3d body pose from silhouettes using activity manifold
  learning.
\newblock In {\em Proc. CVPR}, pages 681--688, 2004.

\bibitem{vunet}
Patrick Esser, Ekaterina Sutter, and Bj{\"{o}}rn Ommer.
\newblock A variational u-net for conditional appearance and shape generation.
\newblock {\em CoRR}, abs/1804.04694, 2018.

\bibitem{goodfellow2014generative}
Ian Goodfellow, Jean Pouget-Abadie, Mehdi Mirza, Bing Xu, David Warde-Farley,
  Sherjil Ozair, Aaron Courville, and Yoshua Bengio.
\newblock Generative adversarial nets.
\newblock In {\em Proc. NIPS}, pages 2672--2680, 2014.

\bibitem{guler2018densepose}
R{\i}za~Alp G{\"u}ler, Natalia Neverova, and Iasonas Kokkinos.
\newblock Densepose: Dense human pose estimation in the wild.
\newblock {\em arXiv preprint arXiv:1802.00434}, 2018.

\bibitem{viton}
Xintong Han, Zuxuan Wu, Zhe Wu, Ruichi Yu, and Larry~S. Davis.
\newblock {VITON:} an image-based virtual try-on network.
\newblock {\em CoRR}, abs/1711.08447, 2017.

\bibitem{resnet}
Kaiming He, Xiangyu Zhang, Shaoqing Ren, and Jian Sun.
\newblock Deep residual learning for image recognition.
\newblock In {\em Proc. CVPR}, pages 770--778, 2016.

\bibitem{dropout}
Geoffrey~E. Hinton, Nitish Srivastava, Alex Krizhevsky, Ilya Sutskever, and
  Ruslan Salakhutdinov.
\newblock Improving neural networks by preventing co-adaptation of feature
  detectors.
\newblock {\em CoRR}, abs/1207.0580, 2012.

\bibitem{batchnorm}
Sergey Ioffe and Christian Szegedy.
\newblock Batch normalization: Accelerating deep network training by reducing
  internal covariate shift.
\newblock In {\em Proc. ICML}, pages 448--456, 2015.

\bibitem{pix2pix2017-cvpr}
Phillip Isola, Jun-Yan Zhu, Tinghui Zhou, and Alexei~A Efros.
\newblock Image-to-image translation with conditional adversarial networks.
\newblock In {\em Proc.~CVPR 2017}, 2017.

\bibitem{perceptualloss}
Justin Johnson, Alexandre Alahi, and Li Fei-Fei.
\newblock Perceptual losses for real-time style transfer and super-resolution.
\newblock In {\em Proc. ECCV}, pages 694--711, 2016.

\bibitem{Kingma2014-Adam}
Diederik~P. Kingma and Jimmy Ba.
\newblock Adam: {A} method for stochastic optimization.
\newblock {\em CoRR}, abs/1412.6980, 2014.

\bibitem{vae}
Diederik~P. Kingma and Max Welling.
\newblock Auto-encoding variational bayes.
\newblock {\em CoRR}, abs/1312.6114, 2013.

\bibitem{LassnerPG17}
Christoph Lassner, Gerard Pons{-}Moll, and Peter~V. Gehler.
\newblock A generative model of people in clothing.
\newblock In {\em Proc. ICCV}, pages 853--862, 2017.

\bibitem{SRGAN}
Christian Ledig, Lucas Theis, Ferenc Huszar, Jose Caballero, Andrew Cunningham,
  Alejandro Acosta, Andrew~P. Aitken, Alykhan Tejani, Johannes Totz, Zehan
  Wang, and Wenzhe Shi.
\newblock Photo-realistic single image super-resolution using a generative
  adversarial network.
\newblock In {\em Proc. CVPR}, pages 105--114, 2017.

\bibitem{SSD}
Wei Liu, Dragomir Anguelov, Dumitru Erhan, Christian Szegedy, Scott~E. Reed,
  Cheng{-}Yang Fu, and Alexander~C. Berg.
\newblock {SSD:} single shot multibox detector.
\newblock In {\em Proc. ECCV}, pages 21--37, 2016.

\bibitem{DeepFashion}
Ziwei Liu, Ping Luo, Shi Qiu, Xiaogang Wang, and Xiaoou Tang.
\newblock Deepfashion: Powering robust clothes recognition and retrieval with
  rich annotations.
\newblock In {\em Proc. CVPR}, 2016.

\bibitem{ma2018exemplar}
Liqian Ma, Xu Jia, Stamatios Georgoulis, Tinne Tuytelaars, and Luc Van~Gool.
\newblock Exemplar guided unsupervised image-to-image translation.
\newblock {\em arXiv preprint arXiv:1805.11145}, 2018.

\bibitem{poseguided}
Liqian Ma, Xu Jia, Qianru Sun, Bernt Schiele, Tinne Tuytelaars, and Luc~Van
  Gool.
\newblock Pose guided person image generation.
\newblock In {\em Proc. NIPS}, pages 405--415, 2017.

\bibitem{Disentangled}
Liqian Ma, Qianru Sun, Stamatios Georgoulis, Luc~Van Gool, Bernt Schiele, and
  Mario Fritz.
\newblock Disentangled person image generation.
\newblock {\em CoRR}, abs/1712.02621, 2017.

\bibitem{LeakyReLU}
Andrew~L Maas, Awni~Y Hannun, and Andrew~Y Ng.
\newblock Rectifier nonlinearities improve neural network acoustic models.
\newblock In {\em Proc. ICML}, volume~30, page~3, 2013.

\bibitem{posemanifold1}
Liang Mei, Jingen Liu, Alfred O.~Hero III, and Silvio Savarese.
\newblock Robust object pose estimation via statistical manifold modeling.
\newblock In {\em Proc. ICCV}, pages 967--974, 2011.

\bibitem{cgan}
Mehdi Mirza and Simon Osindero.
\newblock Conditional generative adversarial nets.
\newblock {\em CoRR}, abs/1411.1784, 2014.

\bibitem{ReLU}
Vinod Nair and Geoffrey~E. Hinton.
\newblock Rectified linear units improve restricted boltzmann machines.
\newblock In {\em Proc. ICML}, pages 807--814, 2010.

\bibitem{neverova2018dense}
Natalia Neverova, Riza Alp~Guler, and Iasonas Kokkinos.
\newblock Dense pose transfer.
\newblock {\em arXiv preprint arXiv:1809.01995}, 2018.

\bibitem{ACGAN}
Augustus Odena, Christopher Olah, and Jonathon Shlens.
\newblock Conditional image synthesis with auxiliary classifier gans.
\newblock In {\em Proc. ICML}, pages 2642--2651, 2017.

\bibitem{unsupervisedposetransfer}
Albert Pumarola, Antonio Agudo, Alberto Sanfeliu, and Francesc Moreno-Noguer.
\newblock Unsupervised person image synthesis in arbitrary poses.
\newblock In {\em Proc. CVPR}, pages 8620--8628, 2018.

\bibitem{radford2015dcgan}
Alec Radford, Luke Metz, and Soumith Chintala.
\newblock Unsupervised representation learning with deep convolutional
  generative adversarial networks.
\newblock {\em arXiv preprint arXiv:1511.06434}, 2015.

\bibitem{garmenttransfer}
Amit Raj, Patsorn Sangkloy, Huiwen Chang, James Hays, Duygu Ceylan, and Jingwan
  Lu.
\newblock Swapnet: Image based garment transfer.
\newblock In {\em Proc. ECCV}, pages 679--695, 2018.

\bibitem{ImageNet}
Olga Russakovsky, Jia Deng, Hao Su, Jonathan Krause, Sanjeev Satheesh, Sean Ma,
  Zhiheng Huang, Andrej Karpathy, Aditya Khosla, Michael~S. Bernstein,
  Alexander~C. Berg, and Fei{-}Fei Li.
\newblock Imagenet large scale visual recognition challenge.
\newblock {\em International Journal of Computer Vision}, 115(3):211--252,
  2015.

\bibitem{improvedtrainingGAN}
Tim Salimans, Ian~J. Goodfellow, Wojciech Zaremba, Vicki Cheung, Alec Radford,
  and Xi Chen.
\newblock Improved techniques for training gans.
\newblock In {\em Proc. NIPS}, pages 2226--2234, 2016.

\bibitem{DeformableGAN}
Aliaksandr Siarohin, Enver Sangineto, St{\'{e}}phane Lathuili{\`{e}}re, and
  Nicu Sebe.
\newblock Deformable gans for pose-based human image generation.
\newblock {\em CoRR}, abs/1801.00055, 2018.

\bibitem{vgg}
Karen Simonyan and Andrew Zisserman.
\newblock Very deep convolutional networks for large-scale image recognition.
\newblock {\em CoRR}, abs/1409.1556, 2014.

\bibitem{inception}
Christian Szegedy, Vincent Vanhoucke, Sergey Ioffe, Jonathon Shlens, and
  Zbigniew Wojna.
\newblock Rethinking the inception architecture for computer vision.
\newblock In {\em Proc. CVPR}, pages 2818--2826, 2016.

\bibitem{instancenorm}
Dmitry Ulyanov, Andrea Vedaldi, and Victor~S. Lempitsky.
\newblock Instance normalization: The missing ingredient for fast stylization.
\newblock {\em CoRR}, abs/1607.08022, 2016.

\bibitem{videogenerating}
Jacob Walker, Kenneth Marino, Abhinav Gupta, and Martial Hebert.
\newblock The pose knows: Video forecasting by generating pose futures.
\newblock In {\em Proc. ICCV}, pages 3352--3361, 2017.

\bibitem{cpviton}
Bochao Wang, Huabin Zheng, Xiaodan Liang, Yimin Chen, Liang Lin, and Meng Yang.
\newblock Toward characteristic-preserving image-based virtual try-on network.
\newblock In {\em Proc. ECCV}, pages 607--623, 2018.

\bibitem{SSIM}
Zhou Wang, Alan~C. Bovik, Hamid~R. Sheikh, and Eero~P. Simoncelli.
\newblock Image quality assessment: from error visibility to structural
  similarity.
\newblock {\em {IEEE} Trans. Image Processing}, 13(4):600--612, 2004.

\bibitem{PCK}
Yi Yang and Deva Ramanan.
\newblock Articulated human detection with flexible mixtures of parts.
\newblock {\em {IEEE} Trans. Pattern Anal. Mach. Intell.}, 35(12):2878--2890,
  2013.

\bibitem{yu2018free}
Jiahui Yu, Zhe Lin, Jimei Yang, Xiaohui Shen, Xin Lu, and Thomas~S Huang.
\newblock Free-form image inpainting with gated convolution.
\newblock {\em arXiv preprint arXiv:1806.03589}, 2018.

\bibitem{humanappearancetransfer}
Mihai Zanfir, Alin-Ionut Popa, Andrei Zanfir, and Cristian Sminchisescu.
\newblock Human appearance transfer.
\newblock In {\em Proc. CVPR}, pages 5391--5399, 2018.

\bibitem{multiview}
Bo Zhao, Xiao Wu, Zhi{-}Qi Cheng, Hao Liu, and Jiashi Feng.
\newblock Multi-view image generation from a single-view.
\newblock {\em CoRR}, abs/1704.04886, 2017.

\bibitem{Market1501}
Liang Zheng, Liyue Shen, Lu Tian, Shengjin Wang, Jingdong Wang, and Qi Tian.
\newblock Scalable person re-identification: {A} benchmark.
\newblock In {\em Proc. ICCV}, pages 1116--1124, 2015.

\bibitem{personreid}
Liang Zheng, Yi Yang, and Alexander~G. Hauptmann.
\newblock Person re-identification: Past, present and future.
\newblock {\em CoRR}, abs/1610.02984, 2016.

\bibitem{unlabeledReID}
Zhedong Zheng, Liang Zheng, and Yi Yang.
\newblock Unlabeled samples generated by {GAN} improve the person
  re-identification baseline in vitro.
\newblock In {\em Proc. ICCV}, pages 3774--3782, 2017.

\bibitem{CycleGANICCV}
Jun{-}Yan Zhu, Taesung Park, Phillip Isola, and Alexei~A. Efros.
\newblock Unpaired image-to-image translation using cycle-consistent
  adversarial networks.
\newblock In {\em Proc. ICCV}, pages 2242--2251, 2017.

\end{thebibliography}
	}
    \clearpage
    
    \begin{center}
    	\Large \textbf{Supplementary materials}
    \end{center}
    \appendix
    
    	\section{Video demonstration}
    We believe a video is more convincing than words and images when evaluating the performance of our method. In this section, we make a video performing some human actions with the appointed appearances sampled from the DeepFashion dataset. The actions are acquired from two sources: 1) clipped dancing videos from the Internet; 2) videos photographed by ourselves. We take three steps to complete such task described as follows. First, we split the videos into sequences of frames and estimate the body joints from those frames through HPE \cite{HPE}. Second, we randomly select several person images as the condition image $P_c$ and regard the afore-estimated joints as target poses $S_t$, and thus generate images frame by frame. Third, we stack the generated images as the original order in the source video and transfer the stacked images into a video. The model we used to generate the video is the combination of our Full method and a revised version of the deformable skip-connections introduced in \cite{DeformableGAN}. We add the deformable skip-connections between the features in the encoder and the features with the same spatial size in the decoder, architecturally similar to the U-Net. Adding these connections can enhance the clothes details at the basis of our APTN, which cooperates to maintain high shape and appearance consistency. The first row in the video represents the condition images. And the first column in the left side gives the source videos, representing the target poses. The demonstration video clearly shows the capability of our method in maintaining appearance and shape consistency. The video is uploaded to: \url{https://youtu.be/bNHFPMX9BVk}.

    \section{Person image generation conditioned on arbitrary poses}
\begin{figure*}
    	\centering
    	\includegraphics[width=\textwidth]{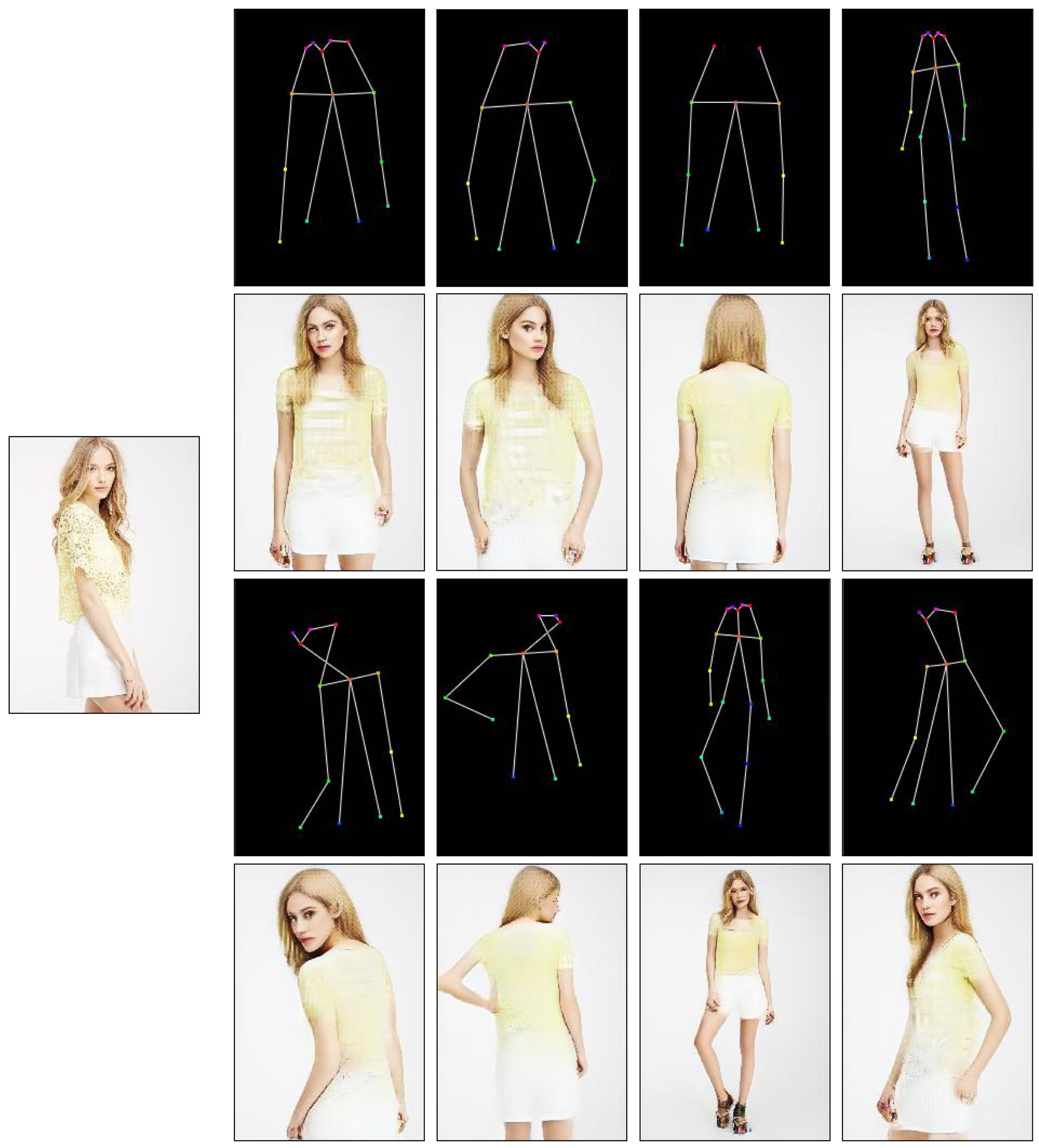}
    	\caption{Illustration of examples (DeepFashion) generated by our method conditioned on different target poses. Note the target poses are selected from the whole test set.}
    	\label{fig:multi_pose_fashion1}
    \end{figure*}

\begin{figure*}
    	\centering
    	\includegraphics[width=\textwidth]{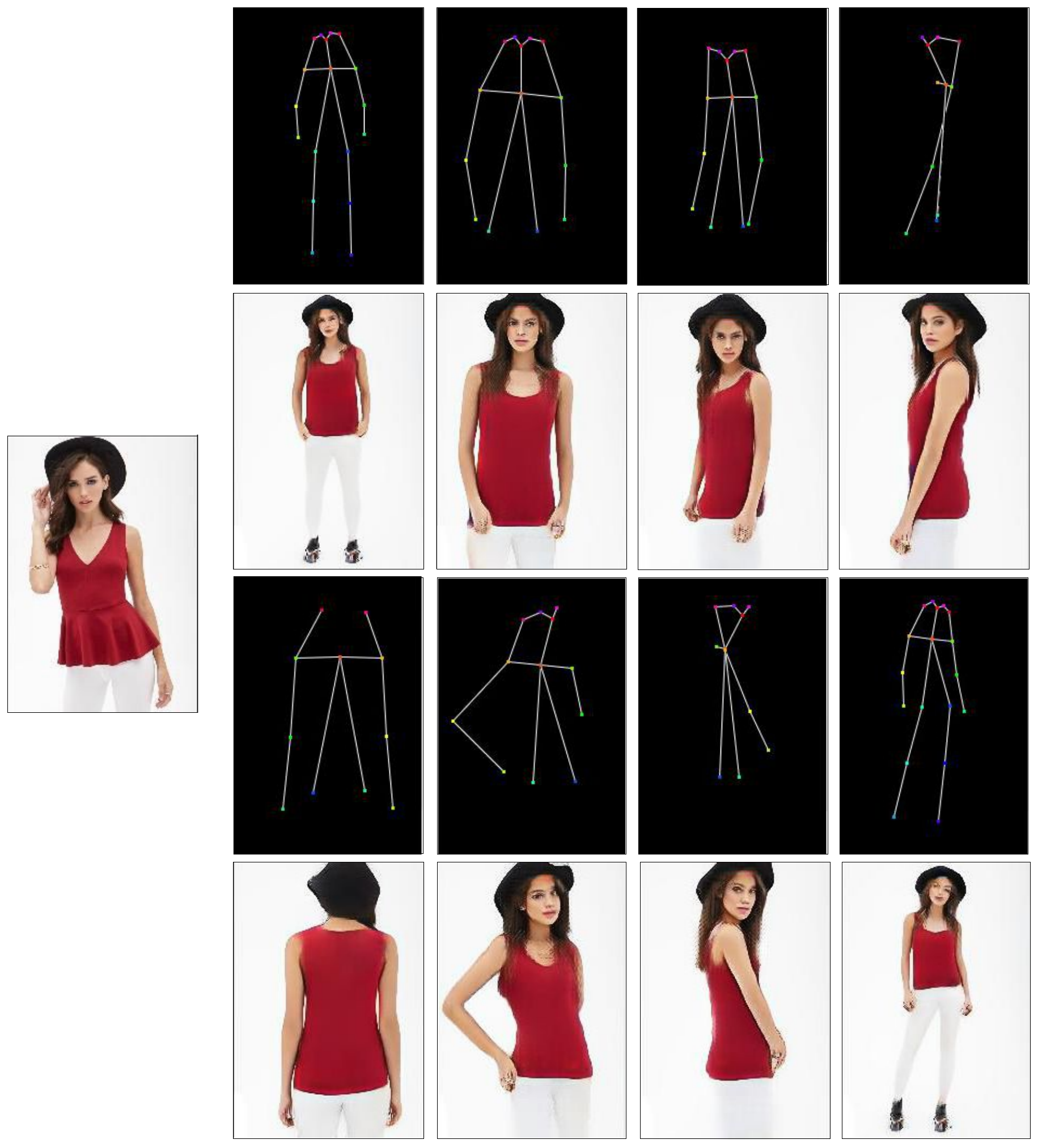}
    	\caption{Illustration of examples (DeepFashion) generated by our method conditioned on different target poses. Note the target poses are selected from the whole test set.}
    	\label{fig:multi_pose_fashion3}
    \end{figure*}

\begin{figure*}
    	\centering
    	\includegraphics[width=\textwidth]{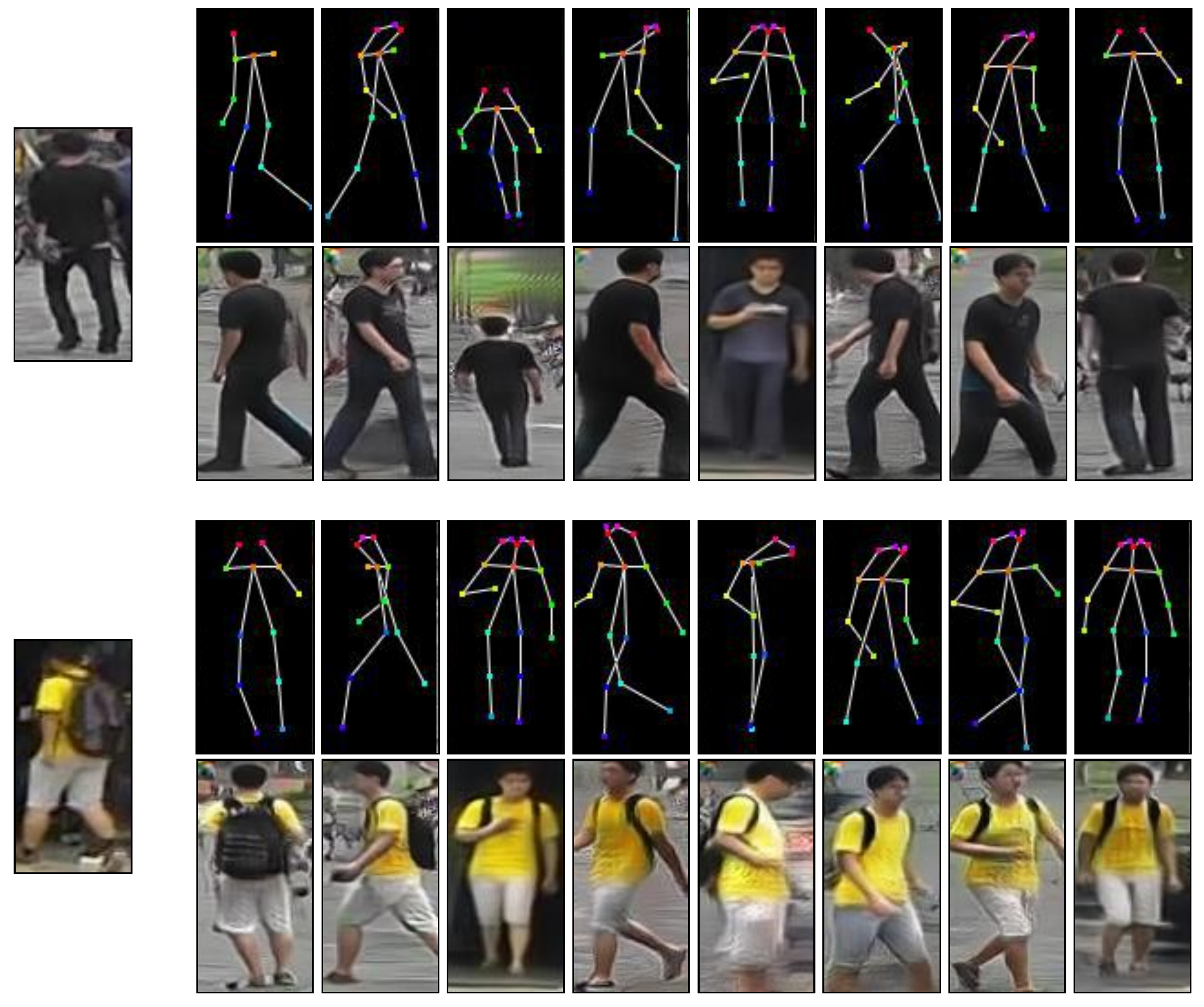}
    	\caption{Illustration of examples (Market-1501) generated by our method conditioned on different target poses. Note the target poses are selected from the whole test set.}
    	\label{fig:multi_pose_market}
    \end{figure*}

    In Fig.\ref{fig:multi_pose_fashion1} and Fig.\ref{fig:multi_pose_fashion3}, we show the person image generation results of our model by giving the same condition person image and several selected poses from the whole test set of DeepFashion dataset. Our model is able to generate natural and realistic results even given poses largely varied in viewpoints, scales and directions etc. In Fig.\ref{fig:multi_pose_market}, we give several generated examples conditioned on the same person image but given different target poses on Market-1501 dataset.
    
    \section{Comparison with previous works}
\begin{figure*}
    	\centering
    	\includegraphics[width=\textwidth]{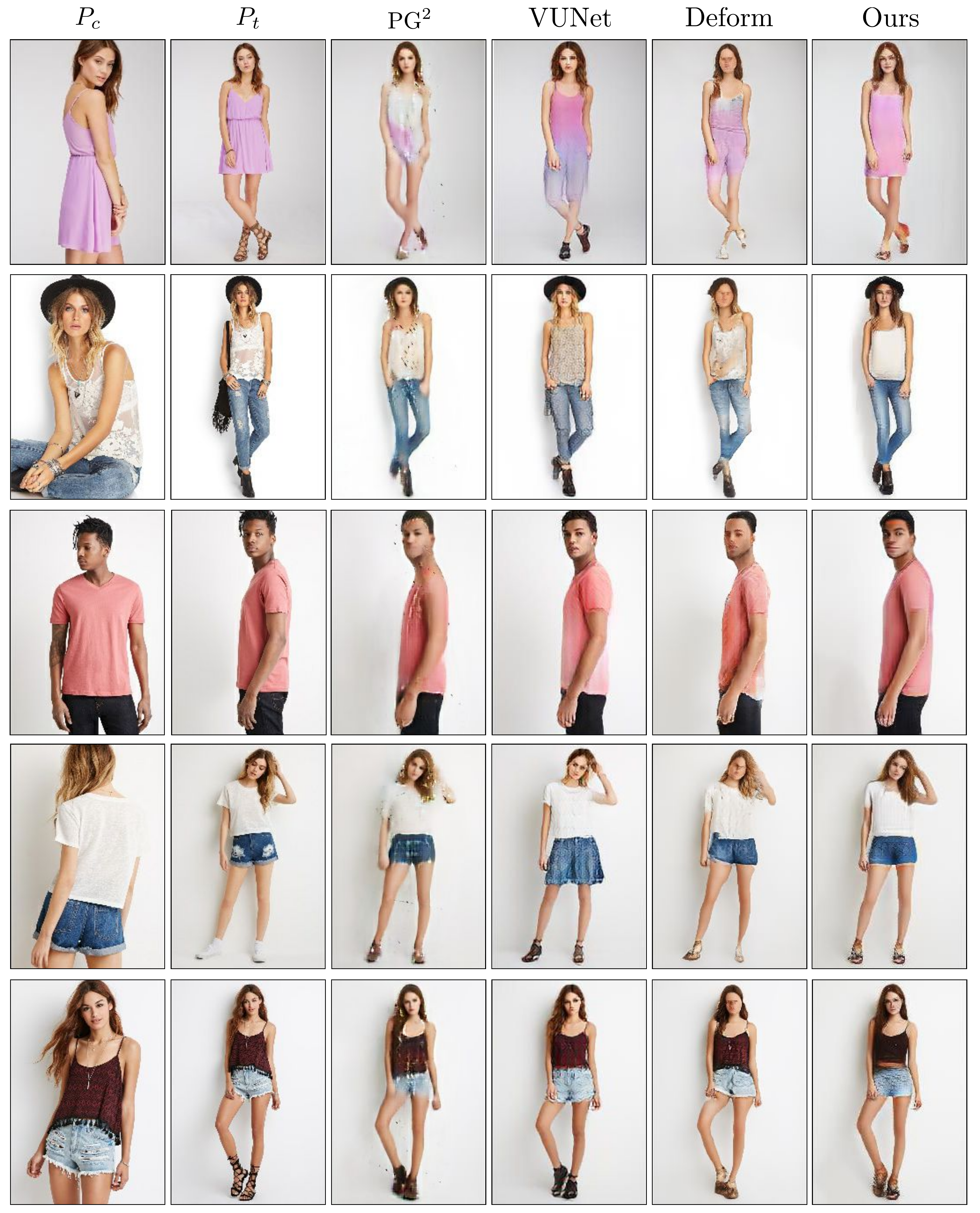}
    	\caption{Qualitative comparisons on DeepFashion Datasets. $\mathrm{PG^2}$, $\mathrm{VUNet}$ and $\mathrm{Deform}$ represent the results obtained from \cite{poseguided} ,\cite{vunet} and \cite{DeformableGAN}, respectively.}
    	\label{fig:comparison on DeepFashion1}
    \end{figure*}

\begin{figure*}
    	\centering
    	\includegraphics[width=\textwidth]{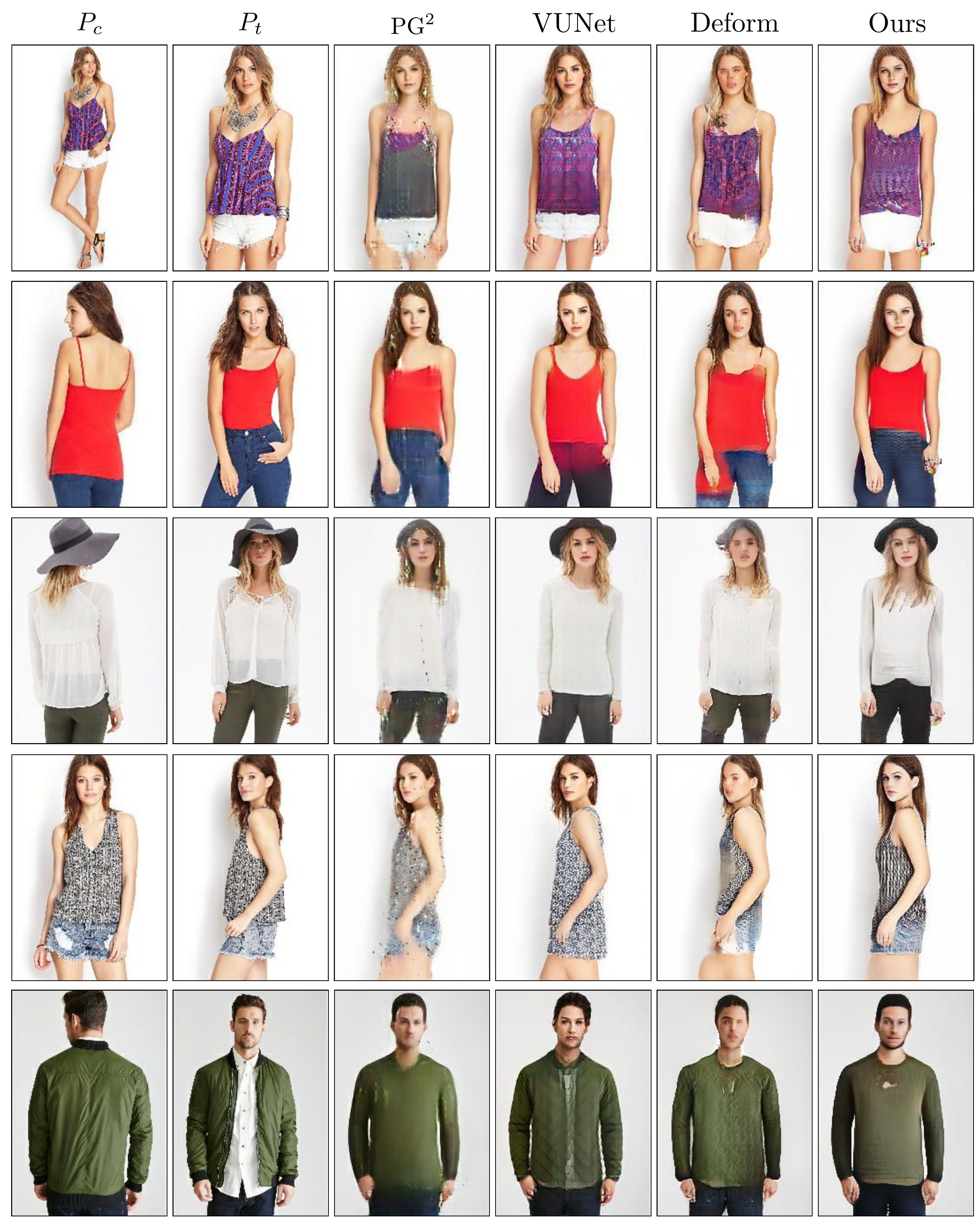}
    	\caption{Qualitative comparisons on DeepFashion Datasets. $\mathrm{PG^2}$, $\mathrm{VUNet}$ and $\mathrm{Deform}$ represent the results obtained from \cite{poseguided} ,\cite{vunet} and \cite{DeformableGAN}, respectively.}
    	\label{fig:comparison on DeepFashion2}
    \end{figure*}

\begin{figure*}
    	\centering
    	\includegraphics[width=0.6\textwidth]{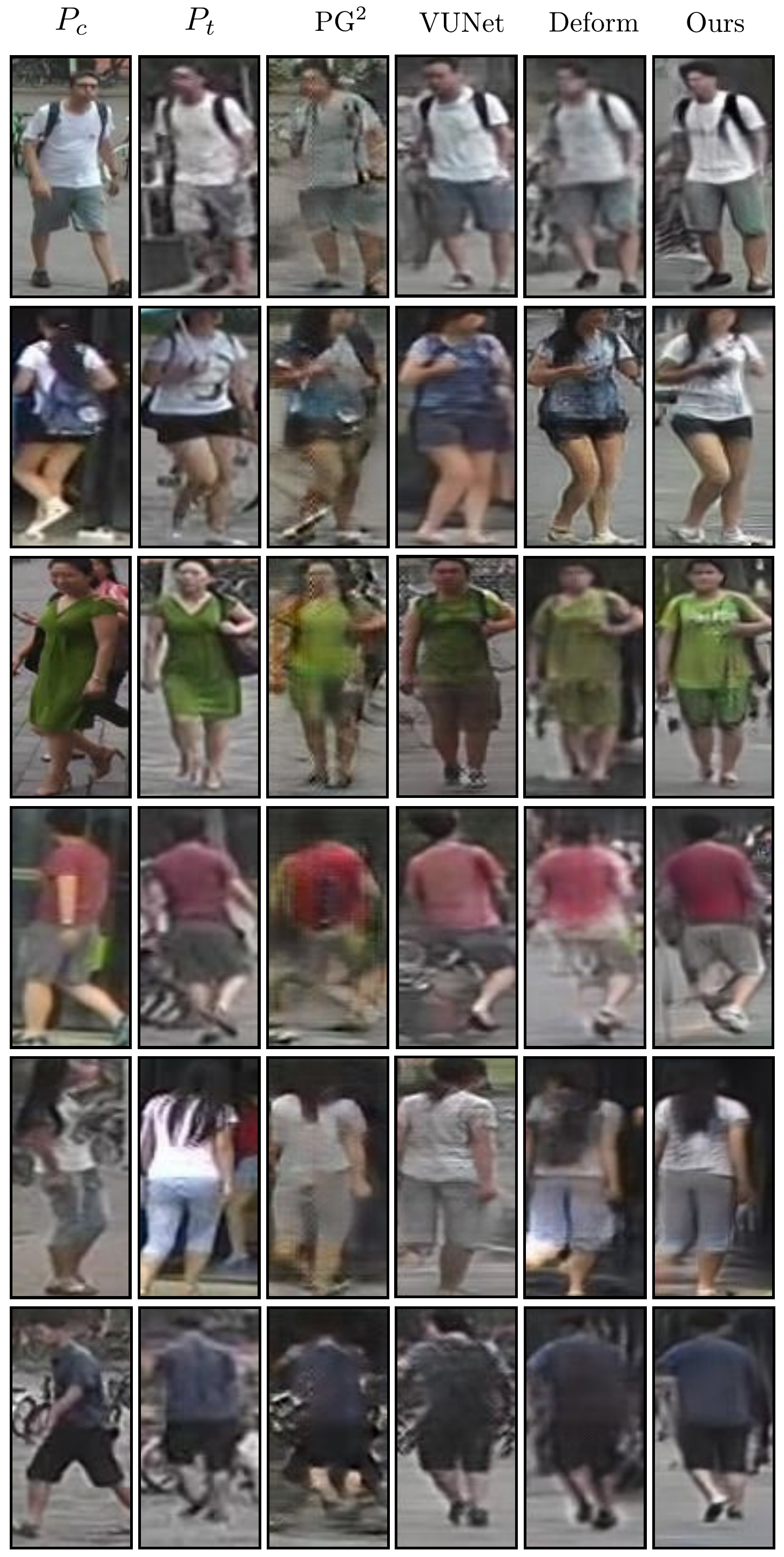}
    	\caption{Qualitative comparisons on Market-1501 Datasets. $\mathrm{PG^2}$, $\mathrm{VUNet}$ and $\mathrm{Deform}$ represent the results obtained from \cite{poseguided} ,\cite{vunet} and \cite{DeformableGAN}, respectively.}
    	\label{fig:comparison on Market1501_1}
    \end{figure*}

\begin{figure*}
    	\centering
    	\includegraphics[width=0.6\textwidth]{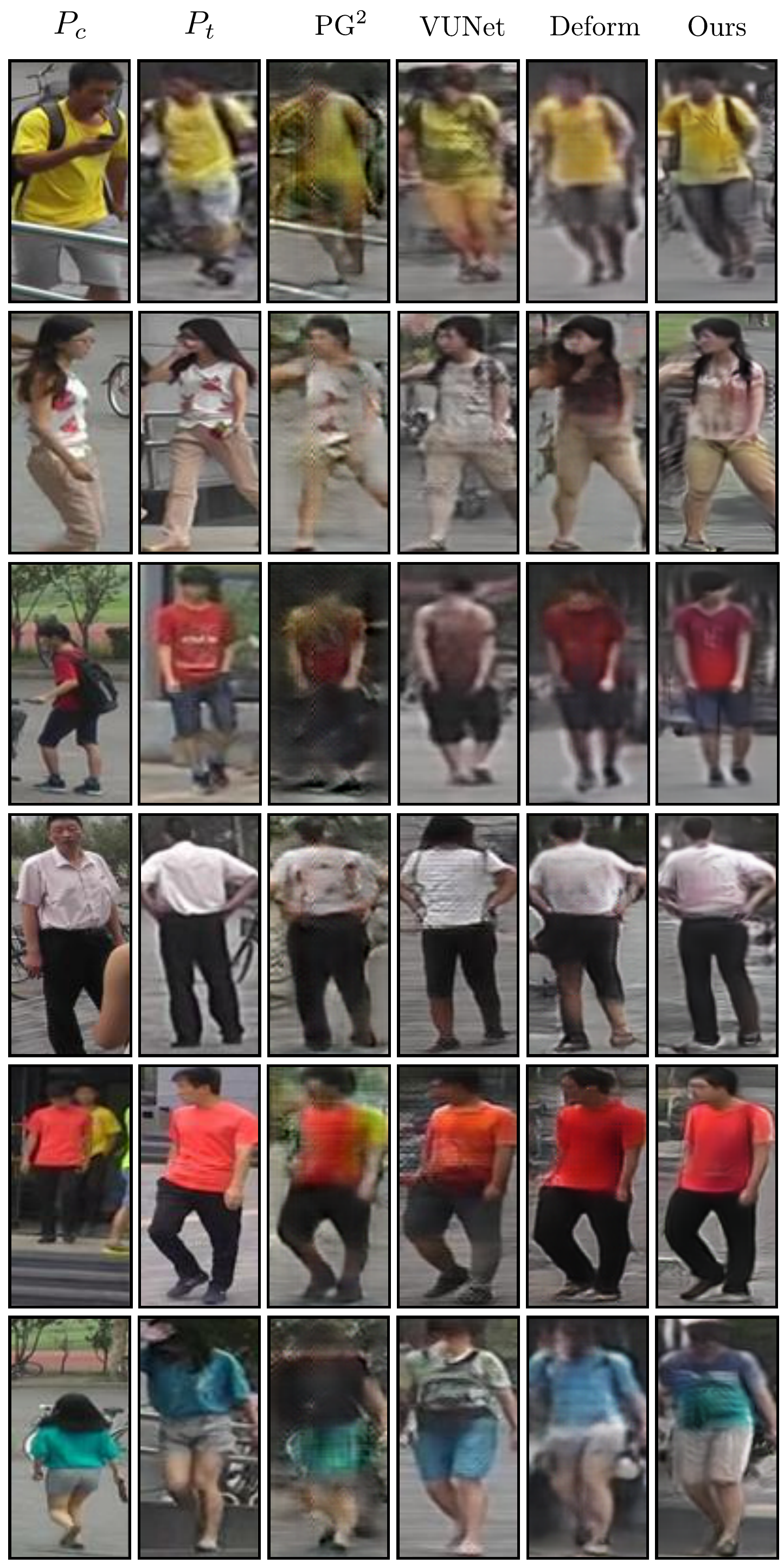}
    	\caption{Qualitative comparisons on Market-1501 Datasets. $\mathrm{PG^2}$, $\mathrm{VUNet}$ and $\mathrm{Deform}$ represent the results obtained from \cite{poseguided} ,\cite{vunet} and \cite{DeformableGAN}, respectively.}
    	\label{fig:comparison on Market1501_2}
    \end{figure*}

    In Fig.\ref{fig:comparison on DeepFashion1} and Fig.\ref{fig:comparison on DeepFashion2}, we give more results of our method (9 PATBs) compared with $\mathrm{PG^2} $\cite{poseguided}, $\mathrm{VUnet}$ \cite{vunet} and $\mathrm{Deform}$ \cite{DeformableGAN} on DeepFashion dataset. In Fig.\ref{fig:comparison on Market1501_1} and Fig.\ref{fig:comparison on Market1501_2}, the comparisons are conducted on Market-1501 dataset. The superiority of our method is clearly shown, mainly reflected on the more consistent appearances and poses with the targets.
    
    \section{More ablation study results}
\begin{figure*}
    	\centering
    	\includegraphics[width=\textwidth]{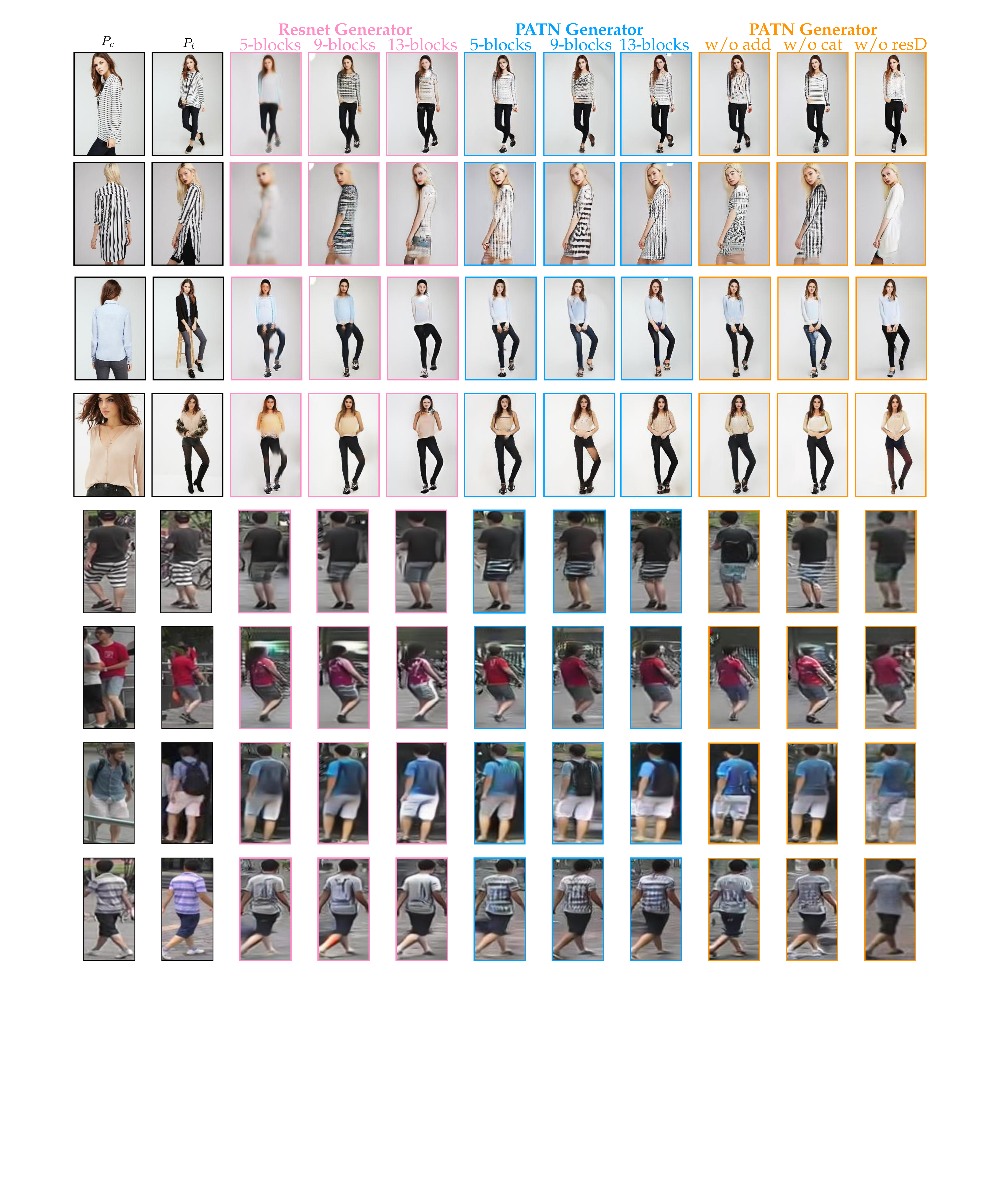}
    	\caption{More qualitative results of ablation study on DeepFashion dataset (Upper) and Market-1501 dataset (Lower).}
    	\label{fig:ablation}
    \end{figure*}

    In Fig.\ref{fig:ablation}, we give more qualitative ablation comparisons on DeepFashion dataset and Market-1501 dataset. These results further demonstrate the advantage of PATN generator over Resnet generator, and at the meantime show that adding more PATBs will make the poses of the generated persons more coherent and natural, revealing the benefits of our progressive generation process.
    
    \section{More visualization examples of the attention masks in PATBs}
\begin{figure*}
    	\centering
    	\includegraphics[width=\textwidth]{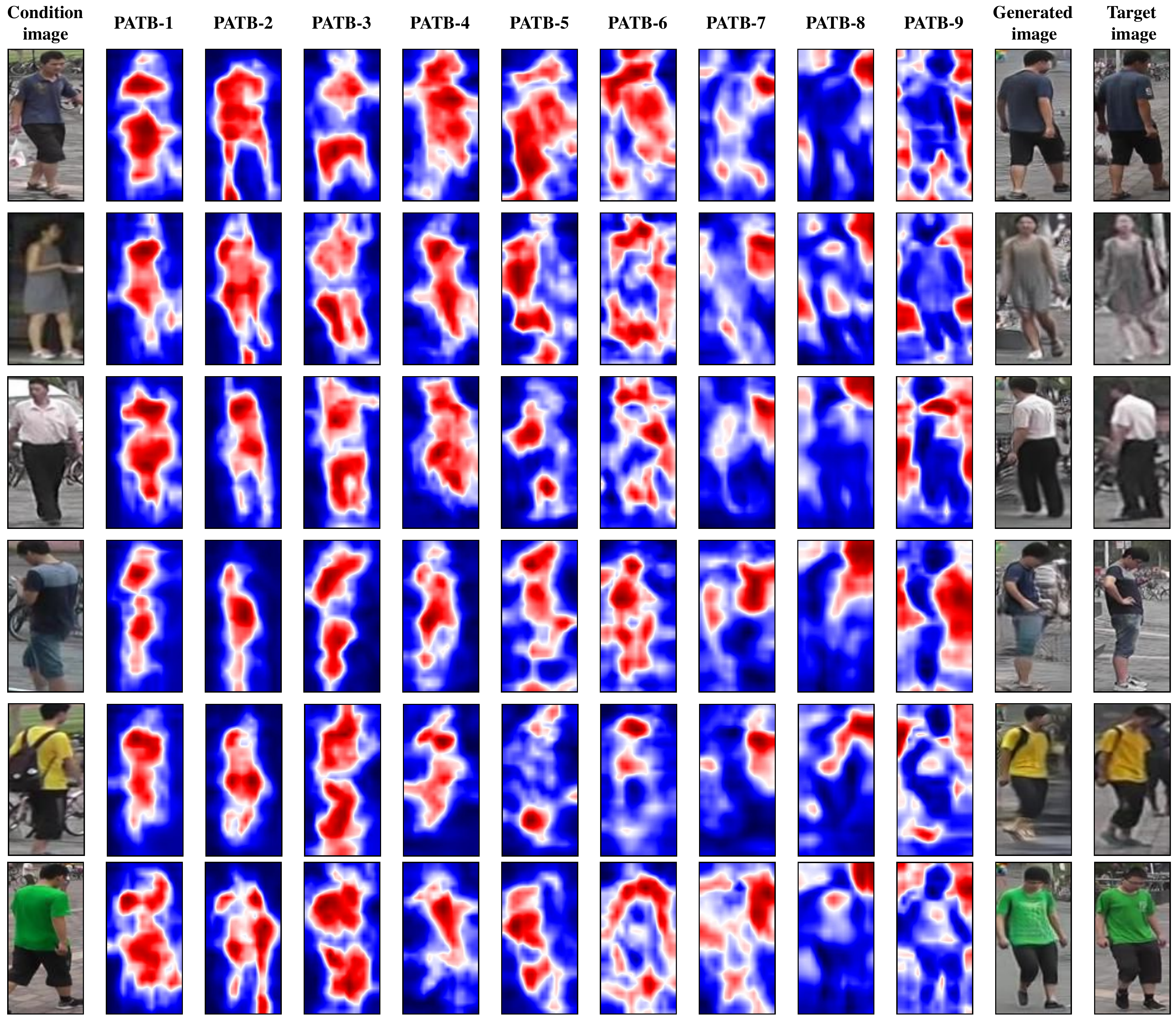}
    	\caption{Visualization of the attention masks $M_t$. ``PATB-1'' denotes the attention mask of the first PATB and ``PATB-2$\sim$9'' likewise.}
    	\label{fig:PAT visualization on market}
    \end{figure*}

    In Fig.\ref{fig:PAT visualization on market}, we give more examples about the visualization of attention masks in PATBs. These results are obtained when our PATN generator with 9 PATBs is tested on Market-1501 dataset. We believe that these extended examples will give an intuition about how the attention mechanism works during the generation process. 
    
    \begin{figure*}
    	\centering
    	\includegraphics[width=0.8\textwidth]{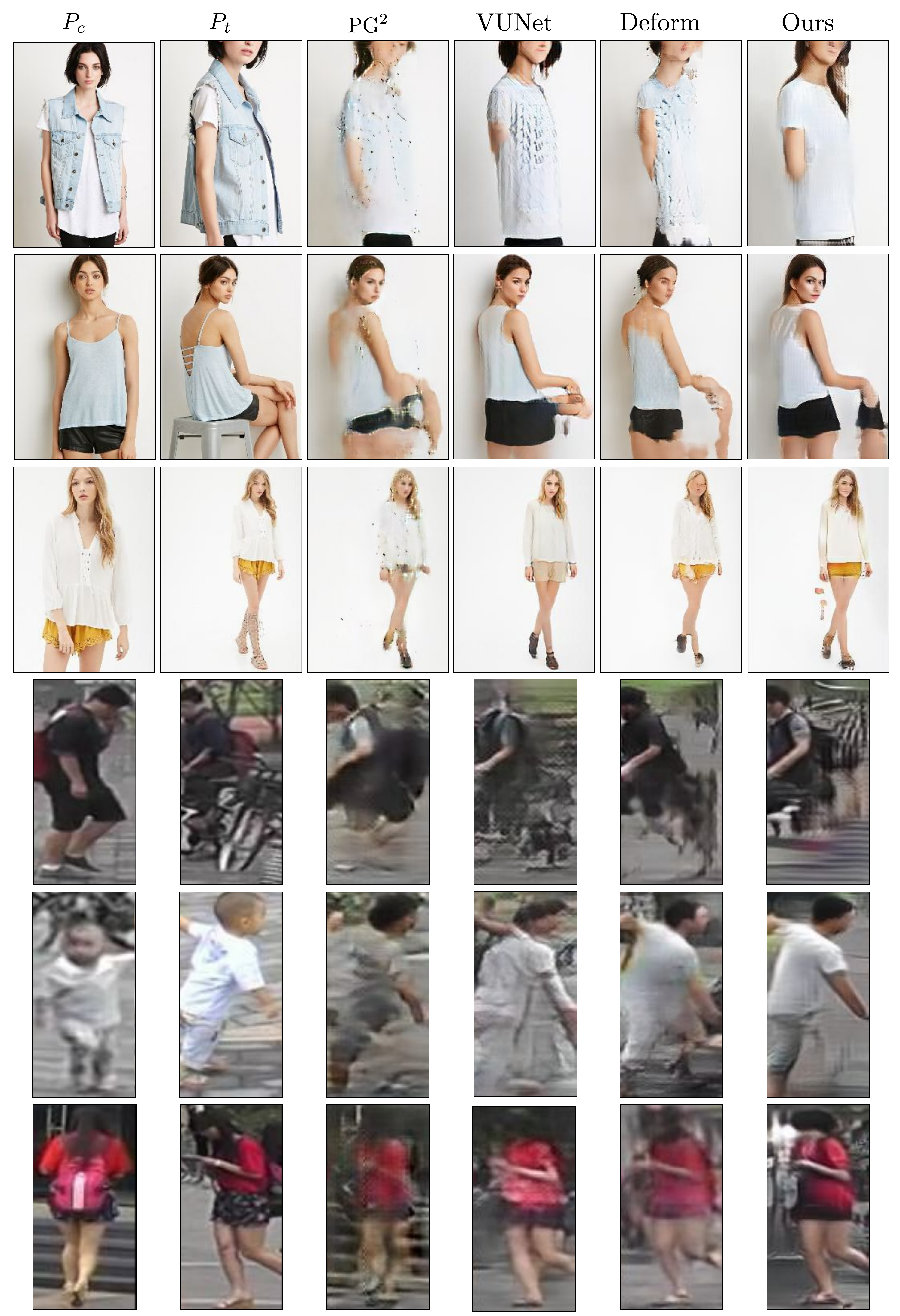}
    	\caption{Some failure cases on DeepFashion (first three rows) and Market-1501 dataset (the left three rows). $\mathrm{PG^2}$, $\mathrm{VUNet}$ and $\mathrm{Deform}$ represent the results obtained from \cite{poseguided} ,\cite{vunet} and \cite{DeformableGAN}, respectively.}
    	\label{fig:failure_cases}
    \end{figure*}
	
\end{document}